\documentclass{article} 
\usepackage{iclr2025_conference,times}

\usepackage{amsmath,amsfonts,bm}

\def\eqref#1{equation~\ref{#1}}

\def\1{\bm{1}}

\DeclareMathAlphabet{\mathsfit}{\encodingdefault}{\sfdefault}{m}{sl}
\SetMathAlphabet{\mathsfit}{bold}{\encodingdefault}{\sfdefault}{bx}{n}

\usepackage{hyperref}
\usepackage{url}

\usepackage{graphicx}
\usepackage{amsmath}
\usepackage{amssymb}
\usepackage{booktabs}
\usepackage{multirow}

\usepackage[capitalize]{cleveref}
\crefname{section}{Sec.}{Secs.}
\Crefname{section}{Section}{Sections}
\Crefname{table}{Table}{Tables}
\crefname{table}{Tab.}{Tabs.}

\title{MagicMirror: A Large-Scale Dataset and Benchmark for Fine-Grained Artifacts Assessment in Text-to-Image Generation}

\author{
Jia Wang$^{1,2}$ \quad Jie Hu$^{2}$\thanks{Corresponding author.} \quad Xiaoqi Ma$^{2}$ \quad Hanghang Ma$^{2}$ \quad Yanbing Zeng$^{2}$ \quad Xiaoming Wei$^{2}$  \\
$^{1}$University of Chinese Academy of Sciences \quad $^{2}$Meituan \quad \\
\{wangj.infinite, hujiemr\}@gmail.com
}

\iclrfinalcopy 
\begin{document}

\maketitle

\begin{abstract}
Text-to-image (T2I) generation has achieved remarkable progress in instruction following and aesthetics. However, a persistent challenge is the prevalence of physical artifacts, such as anatomical and structural flaws, which severely degrade perceptual quality  and limit application.
Given the diversity and complexity of these artifacts, a systematic and fine-grained evaluation framework is required, which is lacking in current benchmarks.
To fill this gap, we introduce \textbf{MagicMirror}, a comprehensive framework for artifacts assessment. 
We first establish a detailed taxonomy of generated image artifacts. Guided by this taxonomy, we manually annotate \textbf{MagicData340K}, the first human-annotated large-scale dataset of 340K generated images with fine-grained artifact labels. Building on this dataset, we train \textbf{MagicAssessor}, a Vision-Language Model (VLM) that provides detailed assessments and corresponding labels. To overcome challenges like class imbalance and reward hacking, we design a novel data sampling strategy and a multi-level reward system for Group Relative Policy Optimization (GRPO). Finally, we leverage MagicAssessor to construct \textbf{MagicBench}, an automated benchmark for evaluating the image artifacts of current T2I models. Our evaluation with MagicBench reveals that despite their widespread adoption, even top-tier models like GPT-image-1 are consistently plagued by significant artifacts, highlighting artifact reduction as a critical frontier for future T2I development. Project page: \url{https://wj-inf.github.io/MagicMirror-page/}.
\end{abstract}

\section{Introduction}
\label{sec:intro}

Diffusion-based text-to-image (T2I) models~\citep{saharia2022photorealistic, rombach2022high, podell2023sdxl, blackforestlabs_flux} have achieved remarkable advancements in image quality, instruction following, and aesthetics. These capabilities have unlocked practical applications in specialized domains, from photorealistic portraiture~\citep{midjourney-v6.1} to graphic design~\citep{gao2025postermaker, hu2025dreamposter}. Yet, this progress is often challenged by a persistent and fundamental problem: the generation of physical artifacts~\citep{xu2023imagereward, liang2024rich}. 
From incorrect limb counts~\citep{zhu2024mole} to distorted shapes, these artifacts represent a fundamental challenge to the models' reliability.
They lower the output's visual quality, require repeated manual correction, and ultimately limit the models' widespread use in real-world situations.

The persistence of these artifacts can be attributed to the historical focus of T2I evaluation, which traditionally prioritized two main areas: image quality (measured by text-agnostic metrics like IS~\citep{salimans2016improved} and FID~\citep{heusel2017gans}), and semantic alignment (assessed with instruction-following benchmarks~\citep{ghosh2023geneval, huang2023t2i, hu2024ella}). More recently, aligning with human preferences is gradually taken into consideration~\citep{kirstain2023pick, xu2023imagereward, wu2023human, wu2023humanv2}, with a significant emphasis on aesthetics~\citep{zhou2024uniaa, liao2025humanaesexpert}. However, while human preference evaluations implicitly consider artifacts, this aspect is not a primary focus and is typically captured only indirectly through an annotator's overall score.

In response to this gap, several recent studies have begun to concentrate more specifically on the task of identifying artifacts. For instance, while RichHF~\citep{liang2024rich} introduces plausibility scores, its approach to artifacts is coarse-grained, labeling all defects with undifferentiated dots. This makes it impossible to distinguish between different error types. Furthermore, its dataset's limited scale (RichHF-18K) and reliance on older T2I models (before the original text-image pairs dataset Pick-a-Pic~\citep{kirstain2023pick}) result in a scarcity of positive examples for challenging subjects, such as anatomically correct hands, hindering effective learning. On the other hand, while subsequent works like HEIE~\citep{yang2025heie} and FakeVLM~\citep{wen2025spot} have explored generating textual explanations for defects, their reliance on standard Supervised Fine-Tuning (SFT) is inherently limited by token-level supervision, which can restrict the flexibility and quality of the model's reasoning. Taken together, these limitations highlight a clear need for a more granular, scalable, and robustly trained evaluation framework.

\begin{figure*}[t]
\centering
\includegraphics[width=1.0\textwidth, trim=0cm 22.5cm 0cm 0cm, clip]{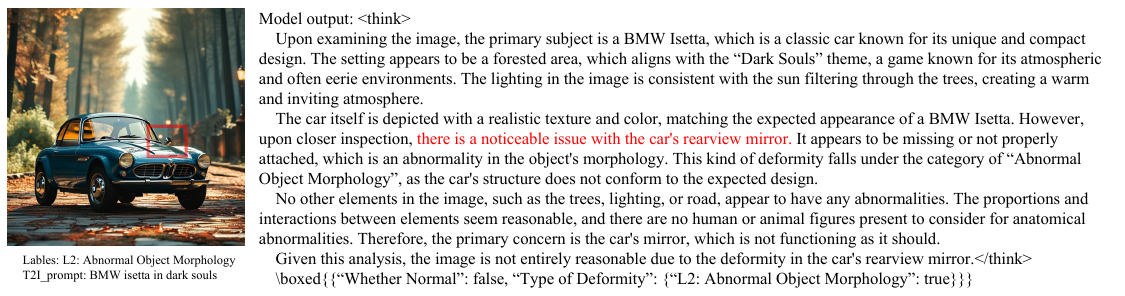} 
\setlength{\abovecaptionskip}{-8pt}
\caption{An output example of MagicAssessor-7B.}
\label{fig:model_response_exp}
\vspace{-13pt}
\end{figure*}

To address this gap, we introduce \textbf{MagicMirror}, a complete framework for systematically evaluating image artifacts. The foundation of our work is a novel, fine-grained taxonomy that categorizes artifacts into three primary groups: \textit{object anatomy}, \textit{attribute}, and \textit{interaction}. 
Guided by this taxonomy, we construct \textbf{MagicData340K}, the first human-annotated large-scale dataset with fine-grained labels in this field. This involves collecting diverse prompts~\citep{kirstain2023pick}, generating approximately 340K images from various advanced T2I models~\citep{esser2024scaling, blackforestlabs_flux, midjourney-v6.1}, and undertaking a massive human annotation effort to apply our detailed artifact labels to each image.

Building upon this data foundation, we introduce \textbf{MagicAssessor}, a specialized Vision-Language Model (VLM) trained specifically for artifact assessment, based on Qwen2.5-VL-7B~\citep{bai2025qwen2}. 
To enhance its performance, we employ Group Relative Policy Optimization (GRPO)~\citep{shao2024deepseekmath}, which we uniquely adapt to our task through two key innovations: a targeted data sampling strategy and a multi-level reward system.
First, to address the issue of data imbalance, our data sampling strategy oversamples challenging positive cases, such as anatomically correct hands. Then, we propose a multi-level reward system, which guides the model from coarse to fine-grained detection and introduce a novel consistency reward to align the model's reasoning with its final output to prevent reward hacking.
We show an output example of MagicAssessor in Fig.~\ref{fig:model_response_exp}, where the artifact of the car's rearview mirror is accurately identified.
Finally, we use this powerful assessor to build \textbf{MagicBench}, our automated benchmark. With MagicBench, we can now fairly compare the image artifacts against various T2I models.
Our main contributions are summarized as follows:

\begin{enumerate}
    \item We develop a comprehensive taxonomy for image artifacts and, guided by it, construct \textbf{MagicData340K}, the first large-scale dataset with fine-grained, human-annotated artifact labels.
    \item We propose \textbf{MagicAssessor}, a fine-grained artifact evaluator, along with a novel training strategy that adapts GRPO with custom data sampling and reward system to overcome challenges like data imbalance and reward hacking.
    \item We build and release \textbf{MagicBench}, the first automated benchmark special for evaluating image artifacts. We leverage it to conduct a systematic analysis of leading T2I models, providing actionable guidance for future work on artifact reduction.
\end{enumerate}

\section{Dataset}

The construction of our dataset follows three main stages. We first establish a fine-grained taxonomy for image artifacts. Guided by this taxonomy, we manually annotate a large collection of text-image pairs generated from diverse and advanced models. Finally, to enable a cold start for our model's step-by-step reasoning, we further annotate a representative subset with localized artifacts and synthesize detailed textual rationales.

\begin{figure*}[t]
\centering
\includegraphics[width=1.0\textwidth, trim=1cm 0cm 1cm 0cm, clip]{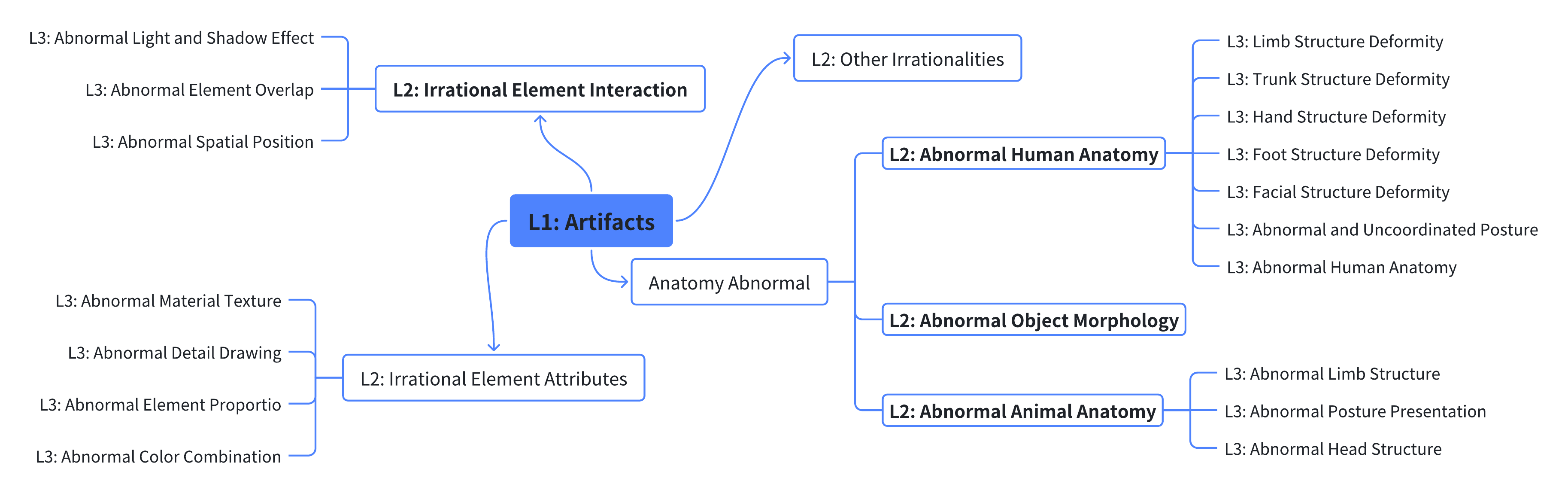} 
\setlength{\abovecaptionskip}{-10pt}
\caption{The hierarchical taxonomy of image artifacts. Our classification begins by distinguishing between artifacts of \textbf{the subject itself} and \textbf{interactions between subjects}. These are further divided into fine-grained Level 2 (L2) and Level 3 (L3) categories.}
\label{fig:artifacts mindmap}
\end{figure*}

\begin{figure*}[t]
\centering
\includegraphics[width=1.0\textwidth, trim=0cm 11.5cm 0.5cm 0cm, clip]{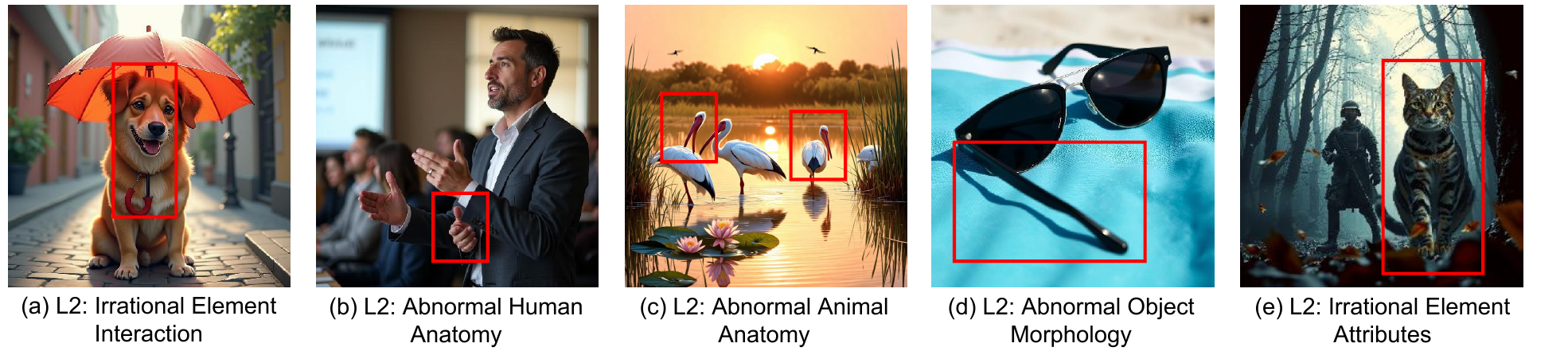} 
\setlength{\abovecaptionskip}{-10pt}
\caption{Visual examples of artifacts corresponding to our taxonomy.}

\label{fig:Four-L2-Label-Example}
\vspace{-10pt}
\end{figure*}

\subsection{Classification of Artifacts}
Existing evaluation methods for generated images lack the necessary granularity. They typically fall into two types: single, coarse-grained scores for metrics like aesthetics or alignment, which offer limited rationale, or undifferentiated spatial annotations like ``artifact region"~\citep{liang2024rich}, which fail to distinguish between different types of artifacts. To enable a truly granular assessment, we propose a multi-label taxonomy that provides a detailed artifact profile for each image.

Our taxonomy is organized hierarchically, as illustrated in Figure~\ref{fig:artifacts mindmap}. We define Normal/Artifact as Level 1 (L1). At the highest level, we distinguish between artifacts concerning the subject itself and those involving interactions between subjects. Subject-level issues are further divided into Anatomy (including human, animal, and object structure) and Attributes (including color and proportion). These main categories constitute our Level 2 (L2) labels. For critical areas, we define more specific Level 3 (L3) labels, such as \textit{Hand Structure Deformity}. Figure~\ref{fig:Four-L2-Label-Example} provides visual examples of several L2 artifacts.

\begin{figure*}[t]
\centering
\includegraphics[width=1.0\textwidth, trim=0cm 4.5cm 2.2cm 0cm, clip]{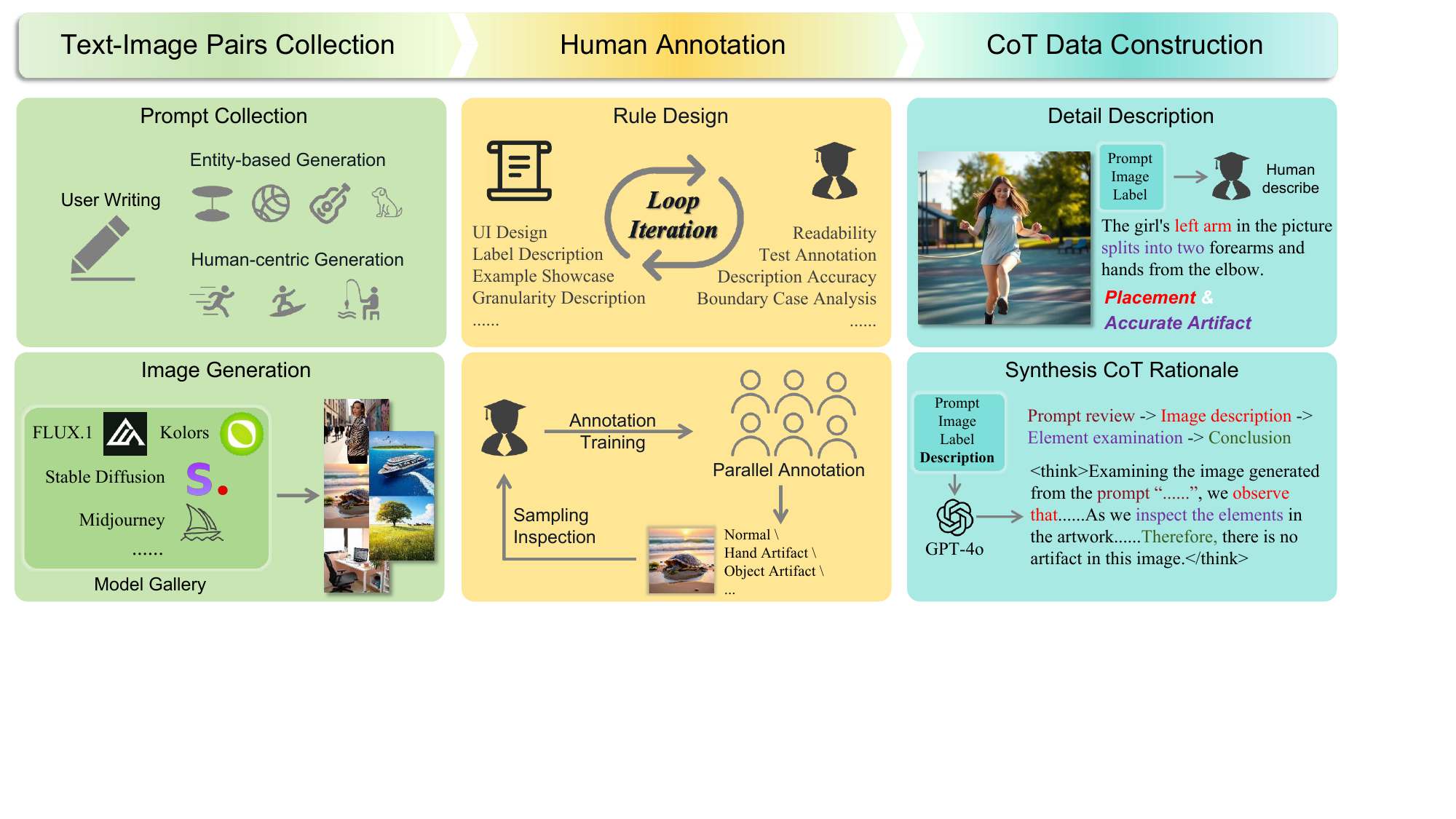} 
\setlength{\abovecaptionskip}{-10pt}
\caption{Data curation for MagicData340K. The process is divided into three main stages: prompt collection and image generation, human annotation with multi-labels, and detailed CoT rationale synthesis for a cold start.}
\label{fig:Data_Curation}
\vspace{-13pt}
\end{figure*}

\subsection{Data Collection}
\label{sec:data collection}
Fig.~\ref{fig:Data_Curation} illustrates the data curation for MagicData340K. The process begins with curating prompts from diverse sources and generating corresponding images with a suite of T2I models. A comprehensive annotation taxonomy for artifacts is then developed through iterative tests and applied to the resulting text-image pairs. Finally, a representative subset is selected for fine-grained annotation, where detailed descriptions for each label are written by humans and used to synthesize Chain-of-Thought (CoT) rationales with GPT-4o.

\textbf{Collecting Text-Image Pairs.}
To construct a diverse dataset, we begin by compiling a large-scale database of entities, artistic styles, and human attributes from a wide range of sources. We then curate a set of 50,000 prompts from three primary sources: (1) 23,000 user prompts sampled from Pick-a-Pic~\citep{kirstain2023pick}; (2) 23,000 prompts generated by GPT combining entities and artistic styles from our database; and (3) 4,000 prompts specifically targeting human subjects, also composed by GPT using attributes from this resource.
We then use this prompt set to generate images with a diverse suite of T2I models, including FLUX.1-dev/schnell~\citep{blackforestlabs_flux}, Kolors1.0~\citep{kolors}, SD3.5~\citep{stabilityai_sd3.5}, SD3~\citep{esser2024scaling}, Midjourney-v6.1~\citep{midjourney-v6.1}, and an internal model.

\textbf{Human Annotation with Multi-Labels.}
After collecting numerous text-image pairs, we start a multi-stage annotation process. We begin by developing a detailed set of annotation guidelines and an intuitive interface, which are refined through a pilot study with expert annotators and several cycles of feedback. The final guidelines provide clear definitions and visual examples for each artifact label, specify the required level of detail, and outline how to handle ambiguous cases (see Appendix~\ref{ap-sec: dataset and label} for details).
With these robust guidelines, our annotators begin the large-scale labeling process. To ensure high quality and consistency, experts continuously follow the annotation work to ensure all rules are followed. Finally, we filter out inappropriate content, resulting in 343,269 annotated text-image pairs. Note that our annotation scheme is designed to assess complex images that may contain multiple artifacts. While the L1 label is a simple binary choice (Normal vs. Artifact), annotators can assign multiple L2 and L3 labels to a single image to describe all co-occurring artifacts.

\textbf{Chain-of-Thought Data Construction.}
To train a model capable of not just detecting but also explaining image artifacts, we curate a high-quality subset from our annotated data for more detailed annotation. For each sample in this subset, human annotators provide detailed textual descriptions for every applied artifact label, specifying the location and nature of the issue.
After gathering this granular information, we feed all data points for a single sample: the original prompt, the generated image, the artifact labels, and their detailed descriptions into GPT-4o. The model is then prompted to synthesize this information into a high-quality, step-by-step CoT rationale. Crucially, during this process, any samples where the initial labels are inaccurate or ambiguous were discarded. This process ensures the detailed subset maintains high quality, making it suitable for fine-tuning models on detailed reasoning tasks.

To acquire high-quality CoT data from GPT-4o, we engineer our prompt using two key strategies. We first prompt the model to follow a structured, four-step reasoning process: \textit{T2I prompt review}, \textit{image description}, \textit{element examination}, and \textit{conclusion} as shown in Fig.~\ref{fig:Data_Curation}. Additionally, we employ an in-context learning (ICL) approach by including a curated example directly within the prompt to demonstrate the desired output format and reasoning style.

\begin{figure*}[t]
\centering
\includegraphics[width=1.0\textwidth, trim=3cm 10cm 0.5cm 1cm, clip]{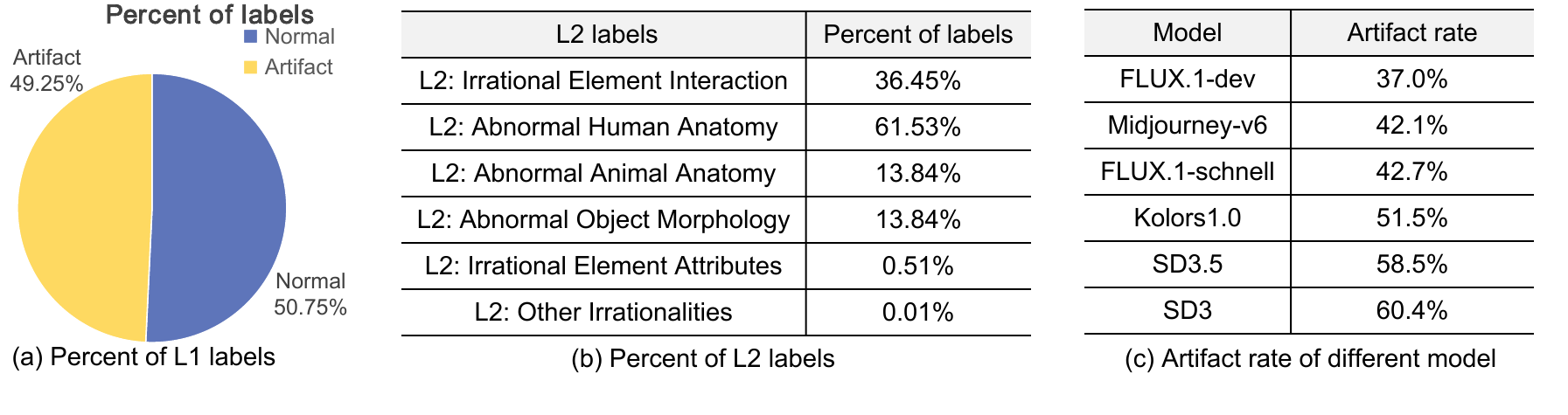} 
\caption{Data statistics of MagicData340K. Note that multiple L2 labels can coexist for a single image, leading the sum of percentages in (b) is over 100\%.}
\setlength{\abovecaptionskip}{-8pt}
\label{fig:Data_Statistics}
\vspace{-10pt}
\end{figure*}

\subsection{Data Statistics of MagicData340K}
Our statistical analysis of the annotated data is presented in Fig.~\ref{fig:Data_Statistics}, which shows a nearly even split between normal images and those containing artifacts. Among the different types of artifacts, \textit{L2: Abnormal Human Anatomy} is the most frequent, caused by the complexity of the human body. Current T2I models often generate images with missing fingers, distorted faces, and unnatural limbs, which are even more common in multi-person scenarios. Another common issue is \textit{L2: Irrational Element Interaction}, where the most frequent sub-type is \textit{L3: Abnormal Element Overlap}. These models struggle to generate clear boundaries between objects, leading to blended or improperly merged images. As for individual model performance, FLUX.1-dev has a relatively low artifact rate, with its outputs showing a notably higher success rate in rendering anatomically correct hands. More details of MagicData340K are shown in Appendix~\ref{ap-sec: dataset and label}.


\section{Methods}

\subsection{Model Training}
The training of our model follows a two-stage pipeline. We begin with a cold start based on our CoT data to teach our model the expected reasoning process and output format. Then, we enhance its detection capabilities using GRPO. To optimize the second stage, we introduce the \textit{Multi-Buckets Data Sampling} strategy to handle class imbalance and the \textit{Multi-level Reward System} to guide the model's learning and mitigate reward hacking.

\textbf{Supervised Fine-Tuning with CoT Data.}
The experiment results in Table.~\ref{table:main result} reveal that existing open-source VLMs exhibit poor zero-shot abilities on the artifact detection task. This weakness makes the direct usage of GRPO ineffective: the algorithm works by refining a model's existing reasoning, but naive open-source VLMs struggle to locate obvious artifacts in an image.
To solve this, we resort to a cold start strategy~\citep{guo2025deepseek}, which is front-loaded by first performing SFT on the VLM with minimal iterations using our pre-constructed CoT data. Through this process, the model learns how to generate responses that include a CoT analysis, how to follow our predefined output format and label taxonomy and, crucially, acquire the fundamental ability to assess artifacts on generated images.

\textbf{Artifacts Recognition Enhancement via GRPO.}
Following cold-start, we transition from token-level supervision to sequence-level optimization to enhance the model's reasoning and detection accuracy. For this, we employ GRPO, whose objective is to refine the policy model $\pi_{\theta}$ by rewarding entire generated sequences that are superior to the average quality within a sampled group. For each question $q$, the optimization objective is formulated as:
\begin{equation}
\begin{array}{c}
    \mathcal{J}_{G R P O}(\theta)=\mathbb{E}_{\left[q \sim P(Q),\left\{o_{i}\right\}_{i=1}^{G} \sim \pi_{\theta_{o l d}}(O \mid q)\right]} \\
    \frac{1}{G} \sum_{i=1}^{G}\left(\min \left(\frac{\pi_{\theta}\left(o_{i} \mid q\right)}{\pi_{\theta_{o l d}}\left(o_{i} \mid q\right)} A_{i}, \operatorname{clip}\left(\frac{\pi_{\theta}\left(o_{i} \mid q\right)}{\pi_{\theta_{o l d}}\left(o_{i} \mid q\right)^{\prime}}, 1-\varepsilon, 1+\varepsilon\right) A_{i}\right)-\beta \mathbb{D}_{K L}\left(\pi_{\theta}| | \pi_{r e f}\right)\right),
\end{array}
\end{equation}
where $G$, $\varepsilon$ and $\beta$ are hyper-parameters, $o_i$ is the output from $\pi_{\theta_{old}}$, $\pi_{\theta}$ is the policy model to be optimized, and $A_i$ is the advantage, computed using a group of rewards $\{R_1,R_2,...,R_G\}$ corresponding to the outputs within each group:
\begin{equation}
    A_i=\frac{R_i-mean(\{R_1,R_2,...,R_G\})}{std(\{R_1,R_2,...,R_G\})}.
\end{equation}

To adapt GRPO for artifacts detection, we design a multi-level reward system to compute $R_i$ and a data sampling strategy, which are detailed below.

\begin{figure*}[t]
\centering
\includegraphics[width=1.0\textwidth, trim=0cm 10cm 5cm 0cm, clip]{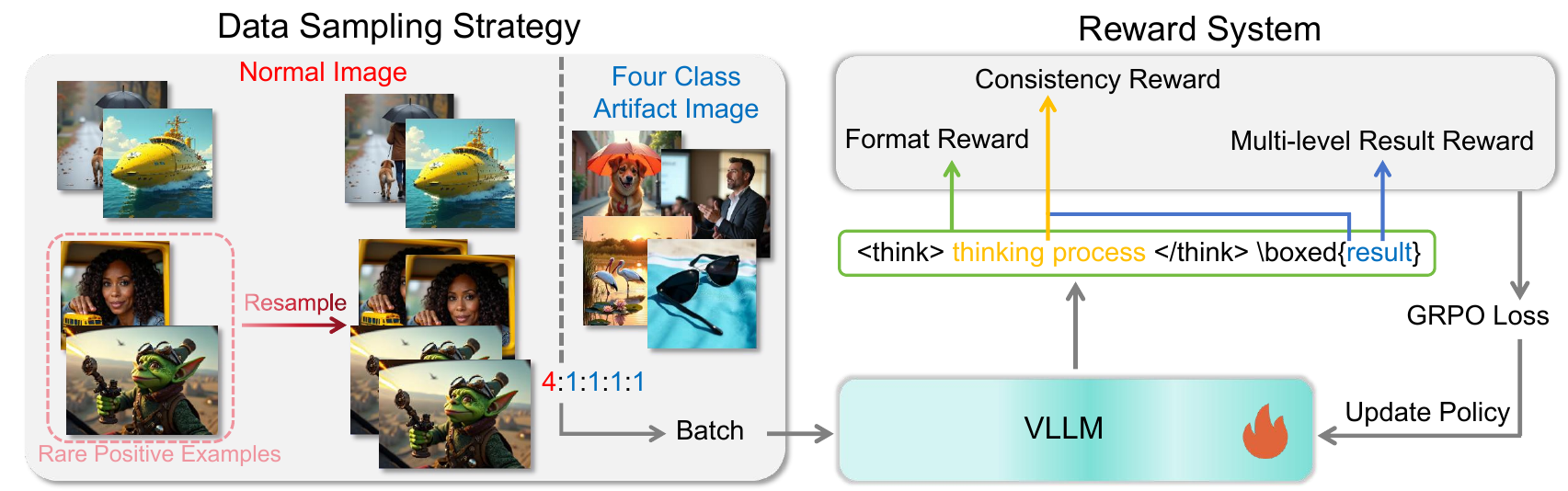} 
\caption{Data sampling strategy and reward system in model training.}
\label{fig:Training_Pipeline}
\vspace{-10pt}
\end{figure*}

\textbf{Multi-level Reward System.}
A simple reward for outcome correctness is insufficient for our complex, hierarchical task. To provide progressive guidance for our model, we design a multi-level reward system. Our design is guided by three core principles: enforcing a structured, hierarchical output format, prioritizing the accuracy of high-level labels over more granular ones, and emphasizing high recall to ensure the model effectively identifies as many artifacts as possible.

Based on these principles, our system computes the final reward $R$, which combines several components. These include a format reward $r_0$, hierarchical rewards for the L1, L2, and L3 labels ($r_1,r_2,r_3$), and a crucial consistency reward $r_c$ that penalizes mismatches between the model's reasoning and its final output to mitigate reward hacking. For the final reward, higher weights are given to more fundamental aspects like format and high-level accuracy. The detailed formulation for each component and the reward calculation are provided in Appendix~\ref{ap-sec: multi-reward}.
The final reward R is a weighted sum:
\begin{equation}
    R=r_c\cdot \sum_{l=0}^{n} 2^{n-l}r_l, n=3,
\end{equation}
where $r_c,r_0,r_1 \in \{0, 1\}$ and $r_2,r_3 \in [0, 1]$.

\textbf{Data Sampling Strategy.}
While cold start provides a promising starting point, the resulting VLM still struggles with the significant class imbalance present in our dataset. A naive application of GRPO would cause the model to overfit to the most frequent artifact categories such as \textit{L2: Abnormal Human Anatomy}, leading to poor recall on rarer types. To address this, we introduce a Multi-Bucket Sampling strategy. For each training batch, we construct a balanced sample by drawing from five distinct buckets: normal images and images corresponding to our four main L2 artifact labels. These are sampled in a 4:1:1:1:1 ratio in a batch, ensuring less frequent but still important artifact types are consistently represented during training. For simplicity, the data of the remaining two L2 labels are omitted from this strategy, which collectively account for less than 1\% of the dataset.

A related challenge is the lack of hard positive samples like anatomically correct human hands, which models typically struggle to generate properly. Without seeing enough of these correct examples, the model may also fall into reward hacking. It could learn a simple but incorrect rule that \textit{any image containing hands is always artifact}. To prevent this, we upsample images with these challenging and anatomically correct objects to better teach the model to distinguish between artifact subjects and correctly formed ones.

\subsection{Assessment Benchmark}
To objectively evaluate and quantify the artifact rates of current T2I models, we introduce MagicBench, a standardized evaluation framework powered by our trained MagicAssessor model.

\textbf{Prompt Construction.} We first construct 800 prompts including single/double/multiple human, single/multiple animals, and single/multiple/complex object (100 for each). Notice that \textit{L2: Irrational Element Interaction} does not have a specific prompt, as it could be present in the generated image containing the entity. To ensure diversity, we use an LLM to generate prompts with different subjects, scenes, photographic styles, and photographic angles, with detail construction process in Appendix~\ref{ap-sec:Prompt Construction of MagicBench}. We also specify in the prompts that the generated image must contain the specified subject. Each model under evaluation is used to generate one image per prompt.

\textbf{Subject Verification.} 
When testing, we found that some models tend to hide the difficult parts of the generated subject (e.g., human hands), or even not include any part of the subject. To address this problem, we highlight it at the end of each prompt and use a general-purpose VLM to perform automated subject verification, confirming that the main subject specified in the prompt is present in the image. 

\textbf{Artifacts Assessment.} 
For all images that successfully pass subject verification, we apply our MagicAssessor for a fine-grained artifact analysis. Our scoring process consists of three steps. 
Images that do not contain the corresponding subject are considered corresponding artifacts and excluded from the subsequent artifacts analysis. After that, we label each generated image with our model. Finally, we give the score according to the model's performance on our test set. The score is calculated by:
\begin{equation}
    Score_{label}=100\cdot(1-\frac{N_{label}}{N_{label\_set}}), 
\end{equation}
where the $N_{label}$ is the image numbers with each L2 label or \textit{Artifacts}, $N_{label\_set}$ is 300 for \textit{Human}, 200 for \textit{Animal}, 300 for \textit{Object}, and 800 for \textit{Interaction} and \textit{Artifacts}. Notice that Overall Score in Table.~\ref{table:benchmark} is the same as $Score_{artifacts}$, representing the performance about whether the generated images are \textit{Normal}.

\section{Experiments}

\begin{table}[]
\caption{Overall performance comparison with other models.}
\label{table:main result}
\centering
\resizebox{0.75\columnwidth}{!}{
\renewcommand{\arraystretch}{1.2}
\begin{tabular}{cccccccccc}
\hline
            & \multicolumn{3}{c}{L2 Labels Macro Average}    & \multicolumn{3}{c}{L2 Labels Micro Average}    & \multicolumn{3}{c}{Artifacts}                       \\ \cmidrule(lr){2-4} \cmidrule(lr){5-7} \cmidrule(lr){8-10} 
                 & Precision       & Recall          & F1              & Precision       & Recall          & F1              & Precision       & Recall          & F1              \\ \cline{1-1} \cline{2-4} \cline{5-7} \cline{8-10}
Qwen2.5-VL-7B    & 0.3395          & 0.0995          & 0.1490          & 0.3263          & 0.1130          & 0.1678          & 0.5674          & 0.3597          & 0.4403          \\
Qwen2.5-VL-32B   & 0.3808          & 0.0289          & 0.0529          & 0.3256          & 0.0301          & 0.0552          & 0.5511          & 0.1320          & 0.2130          \\
InternVL3-8B     & 0.3528          & 0.0958          & 0.1396          & 0.3043          & 0.1054          & 0.1566          & 0.5345          & 0.3617          & 0.4314          \\
InternVL3-38B    & 0.3662          & 0.0313          & 0.0564          & 0.3433          & 0.0282          & 0.0521          & 0.5483          & 0.1034          & 0.1740          \\
GPT-4o           & 0.4348          & 0.1410          & 0.2117          & \underline{0.4923}    & 0.1623          & 0.2442          & \underline{0.6222}    & 0.3189          & 0.4217          \\
Gemini2.5-flash  & \underline{0.4356}    & 0.2753          & 0.3230          & 0.4440          & 0.3068          & 0.3629          & 0.5734          & \underline{0.6682}    & 0.6172          \\
Gemini2.5-pro    & 0.4294          & \underline{0.3572}    & \underline{0.3638}    & 0.3868          & \underline{0.3816}    & \underline{0.3842}    & 0.5577          & \textbf{0.7962} & \underline{0.6560}    \\
\textbf{MagicAssessor-7B} & \textbf{0.5446} & \textbf{0.5244} & \textbf{0.5261} & \textbf{0.5744} & \textbf{0.5425} & \textbf{0.5580} & \textbf{0.7756} & 0.6381          & \textbf{0.7001} \\ \hline
\end{tabular}
}
\vspace{-10pt}
\end{table}

\begin{table}[]
\caption{Detailed performance on L2 labels comparison with other models.}
\label{table:main result 4 class}
\centering
\resizebox{\columnwidth}{!}{
\renewcommand{\arraystretch}{1.2}
\begin{tabular}{ccccccccccccc}
\hline
                   & \multicolumn{3}{c}{L2: Irrational Element Interaction} & \multicolumn{3}{c}{L2: Abnormal Human Anatomy}     & \multicolumn{3}{c}{L2: Abnormal Animal Anatomy}    & \multicolumn{3}{c}{L2: Abnormal Object Morphology}  \\ \cmidrule(lr){2-4} \cmidrule(lr){5-7} \cmidrule(lr){8-10} \cmidrule(lr){11-13} 
                 & Precision         & Recall           & F1               & Precision       & Recall          & F1              & Precision       & Recall          & F1              & Precision       & Recall          & F1              \\ \hline
Qwen2.5-VL-7B    & 0.2008            & 0.1180           & 0.1486           & 0.6182          & 0.1358          & 0.2227          & 0.4361          & 0.1005          & 0.1634          & 0.1028          & 0.0436          & 0.0612          \\
Qwen2.5-VL-32B   & 0.2372            & 0.0296           & 0.0527           & 0.6167          & 0.0331          & 0.0629          & 0.5641          & 0.0337          & 0.0635          & 0.1054          & 0.0193          & 0.0326          \\
InternVL3-8B     & 0.2032            & 0.1509           & 0.1732           & 0.6272          & 0.1065          & 0.1821          & 0.4545          & 0.0798          & 0.1357          & 0.1263          & 0.0459          & 0.0673          \\
InternVL3-38B    & 0.2268            & 0.0383           & 0.0655           & 0.5702          & 0.0230          & 0.0443          & 0.5282          & 0.0566          & 0.1023          & 0.1398          & 0.0071          & 0.0134          \\
GPT-4o           & 0.2612            & 0.1148           & 0.1595           & 0.7193          & 0.2215          & 0.3387          & \underline{0.5563}    & 0.1836          & 0.2761          & \underline{0.2025}    & 0.0440          & 0.0724          \\
Gemini2.5-flash  & \underline{0.2784}      & 0.1288           & 0.1761           & 0.7257          & 0.4235          & 0.5349          & 0.5513          & 0.2608          & 0.3541          & 0.1871          & 0.2881          & 0.2269          \\
Gemini2.5-pro    & 0.2574            & \underline{0.2206}     & \underline{0.2375}     & \underline{0.7523}    & \underline{0.4707}    & \underline{0.5791}    & 0.5475          & \underline{0.3241}    & \underline{0.4072}    & 0.1606          & \textbf{0.4134} & \underline{0.2313}    \\
\textbf{MagicAssessor-7B} & \textbf{0.3665}   & \textbf{0.4621}  & \textbf{0.4088}  & \textbf{0.8957} & \textbf{0.6177} & \textbf{0.7312} & \textbf{0.5805} & \textbf{0.6563} & \textbf{0.6161} & \textbf{0.3359} & \underline{0.3616}    & \textbf{0.3482} \\ \hline
\end{tabular}
}
\vspace{-10pt}
\end{table}

\subsection{Experimental setup}
We develop MagicAssessor by fine-tuning Qwen2.5-VL-7B. Our two-stage training process consists of an initial SFT on the CoT sub-dataset, followed by GRPO on the resampled MagicData340K dataset. We evaluate model performance using Precision, Recall, and F1-Score, calculated for both the overall artifact detection task and for each of our primary L2 categories. All experiments are conducted on 32 NVIDIA H800 GPUs. Comprehensive details regarding our dataset and training hyperparameters are provided in Appendix~\ref{ap-sec:Experimental Details}. In all result tables presented following, we indicate the best-performing value for each metric in \textbf{bold} and the second-best value with \underline{underline}.

\subsection{Performance Analysis}
We perform a comprehensive comparison of MagicAssessor-7B against several baseline models, including Qwen2.5-VL-7B/32B~\citep{bai2025qwen2}, InternVL3-8B/38B~\citep{zhu2025internvl3}, GPT-4o~\citep{achiam2023gpt}, and Gemini2.5-flash/pro~\citep{gemini2024report}. The overall and class-specific performance metrics are detailed in Table~\ref{table:main result} and Table~\ref{table:main result 4 class}, respectively. We present a case of the model's output in Fig.~\ref{fig:model_response_exp} and more cases in Appendix~\ref{ap-sec:Response of Different Models}.

\textbf{Performance of MagicAssessor.}
On the binary classification task, our model achieves a precision of 0.77 and an F1-score of approximately 0.7, indicating its strong potential for use as a reward signal. Breaking down the performance across the four L2 categories, the model excels at identifying human and animal anatomy artifacts but is less effective with interaction and object morphology issues. These latter categories present distinct challenges: for interaction, the model struggles to distinguish between element overlap and image areas with low quality. For object morphology, the sheer diversity and vast number of object types make assessment fundamentally difficult.

\textbf{Comparison with Other Models.}
Our model significantly outperforms all competitors across the main evaluation metrics, establishing a large performance gap. The general-purpose open-source models, like the Qwen-VL and InternVL series, are not very sensitive to artifacts and have very low recall. An unexpected phenomenon is that the larger versions of these models often perform worse. We suggest that larger models are more conservative and tend to regard the images as normal. The commercial models, such as GPT-4o and the Gemini family, perform better but show a clear trade-off. GPT-4o is more precise but tends to be too cautious and misses many actual artifacts. In contrast, the Gemini series is better at finding artifacts but also makes more mistakes, flagging labels that are not real artifacts. In contrast, our model overcomes this trade-off, achieving an excellent balance between finding real issues and not making false claims, which leads to its much better overall performance.

\begin{table}[]
\caption{Evaluation score ($\uparrow$) on different labels of different models in MagicBench.}
\label{table:benchmark}
\centering
\resizebox{\columnwidth}{!}{
\begin{tabular}{cccccc}
\hline 
            & Interaction Score & Human Score     & Animal Score    & Object Score    & Overall Score  \\ \hline
FLUX.1-dev         & \textbf{84.71}             & \textbf{46.00}           & \textbf{44.50}           & 89.60           & \textbf{62.16}           \\
Seedream3.0       & 78.37             & 41.00           & \underline{43.00}           & \textbf{90.56}           & \underline{59.54}           \\
Qwen-image        & 79.01             & \underline{44.15}           & 40.70           & 87.85           & 59.41           \\
Hidream-l1        & 80.30             & 39.33           & 39.50           & \underline{90.07}           & 58.08           \\
FLUX.1-schnell     & 79.90             & 37.67           & 39.00           & 86.49           & 56.03           \\
SD3.5              & 79.06             & 39.33           & 37.50           & 82.64           & 54.82           \\
Kolors1.0         & \underline{82.56}             & 39.46           & 40.50           & 75.44           & 52.82           \\
SD3               & 73.89             & 27.00           & 39.50           & 80.35           & 50.06           \\
SDXL              & 76.02             & 32.33           & 39.00           & 74.30           & 49.36           \\ \hline
GPT-image-1       & \underline{81.54}             & \underline{45.00}           & \textbf{49.00}           & \textbf{91.41}           & \textbf{63.08}           \\
Bagel             & \textbf{85.96}             & 41.00           & \underline{48.00}           & \underline{87.92}           & \underline{60.53}           \\
Blip3-o         & 79.57             & \textbf{46.33}           & 43.00           & 81.27           & 57.98           \\
Janus-pro       & 74.14             & 22.07           & 32.66           & 80.84           & 45.35           \\
Show-o            & 74.49             & 23.33           & 30.50           & 76.41           & 44.77           \\ \hline
\end{tabular}
}
\vspace{-10pt}
\end{table}

\subsection{MagicBench Results}
To evaluate the performance of existing models with respect to image artifacts, we establish a baseline using our MagicBench benchmark. This involves a comprehensive evaluation of leading text-to-image models, with results presented in Table~\ref{table:benchmark}. We select a variety of models, including Qwen-image~\citep{wu2025qwen}, Seedream3.0~\citep{gao2025seedream}, Hidream-l1~\citep{cai2025hidream}, FLUX.1-dev/schell~\citep{blackforestlabs_flux}, Kolors1.0~\citep{kolors}, SD3.5~\citep{stabilityai_sd3.5}, SD3~\citep{esser2024scaling}, SDXL~\citep{podell2023sdxl}, GPT-image-1~\citep{gptimage}, Bagel~\citep{deng2025emerging}, Blip3-o~\citep{chen2025blip3}, Show-o~\citep{xie2024show}, and Janus-pro~\citep{chen2025janus}. These models are divided into two categories for comparison: diffusion-based architectures (e.g., SDXL, FLUX.1-dev) and unified generation-understanding models (e.g., GPT-image-1, Janus-pro).

Among all models, GPT-image-1 and FLUX.1-dev exhibit the highest overall scores. These models demonstrate a strong ability to produce logically sound and anatomically correct images. Conversely, models like Janus-pro and Show-o struggle the most with image artifacts, indicating ongoing difficulties in generating consistently accurate and detailed images.

The result indicates that advanced unified models outperform advanced diffusion-based models on MagicBench. Unified models secure the top position in every evaluation category. Specifically, GPT-image-1 performances best in both Animal Score and Object Score, in addition to achieving the highest Overall Score. Meanwhile, Bagel excels with the best Interaction Score, and Blip3-o leads in Human Score.
While the top diffusion-based models like FLUX.1-dev and Seedream3.0 are highly competitive, they typically rank just behind the leading unified models in most categories. For instance, FLUX.1-dev achieves the highest Overall Score among all diffusion models and performs strongly across the board, but does not secure the top rank in any single category. Similarly, Seedream3.0 is a very close runner-up for the top Object Score. This suggests that the integration of generation and understanding capabilities within a unified architecture may offer an advantage in mitigating the creation of image artifacts.

Consistent with public benchmarks, newer models generally perform better on our benchmark, though FLUX.1-dev notably outperforms the more recent Seedream3.0 and Qwen-image. We attribute this to FLUX.1-dev's tendency to generate subjects in more stable, common poses. In contrast, Seedream3.0 produces more aesthetic and content-rich images, but this complexity increases the likelihood of artifacts. Our findings suggest that for practical application, future model development must balance the pursuit of aesthetics with the critical need for artifact reduction.

\subsection{Ablation Study}

\begin{table}[]
\caption{Overall performance of ablation study.}
\label{table:ablation}
\centering
\resizebox{\columnwidth}{!}{
\renewcommand{\arraystretch}{1.2}
\begin{tabular}{ccccccccccc}
\hline
                                           &                            & \multicolumn{3}{c}{L2 Labels Macro Average}     & \multicolumn{3}{c}{L2 Labels Micro Average}     & \multicolumn{3}{c}{Artifacts}                       \\ \cmidrule(lr){3-5} \cmidrule(lr){6-8} \cmidrule(lr){9-11} 
                                           &                            & Precision       & Recall          & F1              & Precision       & Recall          & F1              & Precision       & Recall          & F1              \\ \hline
\multicolumn{1}{c|}{\multirow{4}{*}{GRPO}} & MagicAssessor-7B           & \underline{ 0.5446}    & 0.5244          & \textbf{0.5261} & \textbf{0.5744} & \underline{ 0.5425}    & \textbf{0.5580} & \textbf{0.7756} & 0.6381          & \underline{ 0.7001}    \\
\multicolumn{1}{c|}{}                      & w/o Multi-Bucket Sampling         & 0.3936          & \textbf{0.5839} & 0.4177          & 0.4497          & \textbf{0.6135} & 0.5190          & 0.5337          & \textbf{0.9652} & 0.6874          \\
\multicolumn{1}{c|}{}                      & w/o Positive Data Resampling & \textbf{0.5462} & 0.5098          & \underline{0.5153}          & \underline{0.5569}          & 0.5196          & \underline{0.5376}          & \underline{0.7725}          & 0.5937          & 0.6714          \\
\multicolumn{1}{c|}{}                      & w/o Consistent Reward      & 0.5292          & \underline{0.5331}          & 0.5120          & 0.5153          & 0.5346          & 0.5248          & 0.7448          & 0.6026          & 0.6662          \\ \hline
\multicolumn{2}{c}{SFT}                                                & 0.4538          & 0.4340          & 0.4306          & 0.5238          & 0.5330          & 0.5284          & 0.7205          & \underline{0.6936}          & \textbf{0.7068} \\ \hline
\end{tabular}
}
\end{table}

We conduct ablation studies to validate our two-stage training pipeline and key design choices, with results in Table~\ref{table:ablation}. Detailed per-class performance metrics are provided in the Appendix (Table~\ref{table:ablation 4 class}). The results show that although MagicAssessor-7B is slightly lower than SFT in the F1 score on the binary classification, it performs significantly better on the four categories of labels. In addition, our model outputs an inspection process, increasing the confidence level of the inspection results.

Our data sampling strategies are crucial for achieving both balanced and precise results. Without \textbf{Multi-Bucket Sampling}, the model overfits to the most frequent artifact type, \textit{L2: Abnormal Human Anatomy}, causing recall on minority classes like \textit{L2: Irrational Element Interaction} to collapse to near zero, as shown in Table~\ref{table:ablation 4 class} in the Appendix. Furthermore, removing \textbf{Positive Data Resampling} degrades the model's ability to distinguish between correctly formed and artifact hands, leading to lower precision on \textit{L2: Abnormal Human Anatomy}. Besides, the design of our reward system is vital for guiding the model effectively and mitigating reward hacking. Removing \textbf{Consistency Reward} leads to a general drop in metrics and more disorganized textual explanations. 

\section{Conclusions and Future Work}
This work introduces MagicMirror, a comprehensive framework to systematically evaluate physical artifacts in visual generation models. It comprises the first human-annotated large-scale dataset in this field (MagicData340K), a specialized assessor (MagicAssessor), and an automated benchmark (MagicBench). Collectively, these tools enable the community to transition from merely acknowledging generation artifacts to actively diagnosing and addressing them. 

While our current framework is designed for post-hoc assessment, a key future direction is to integrate it directly into the model training lifecycle. This can be pursued in two primary ways. First, for static monitoring, it can be used during the pre-training stage to evaluate model checkpoints. In this capacity, it would serve as a diagnostic tool to track convergence with respect to artifact reduction, offering insights into the training process itself. Second, for active optimization, MagicAssessor can be employed as a reward model in a post-training phase, enabling the fine-tuning of T2I models via RLHF or ReFL to directly reduce artifact generation.

\bibliography{iclr2025_conference}
\bibliographystyle{iclr2025_conference}

\newpage

\appendix
\section{Appendix}
This supplementary material is structured into several sections to provide additional details and analysis for our work. Specifically, it covers the following topics:
\begin{itemize}
    \item In Appendix~\ref{ap-sec:related works}, we provide a review of related works in the fields of text-to-image generation and evaluation.
    \item In Appendix~\ref{ap-sec: dataset and label}, we offer a detailed overview of our MagicData340K dataset, including the annotation process and complete label taxonomy.
    \item In Appendix~\ref{ap-sec:prompt design}, we describe the prompt engineering used to elicit detailed reasoning from our models.
    \item In Appendix~\ref{ap-sec: multi-reward}, we elaborate on the design and implementation of the multi-level reward system.
    \item In Appendix~\ref{ap-sec:Experimental Details}, we present the full experimental setup, including training configurations and evaluation metric definitions.
    \item In Appendix~\ref{ap-sec:Response of Different Models}, we showcase qualitative examples of responses from different assessor models on the artifact detection task.
    \item In Appendix~\ref{ap-sec:Prompt Construction of MagicBench}, we present the detail prompt construction process of MagicBench and more detail evaluation result.
\end{itemize}

\subsection{Related Works}
\label{ap-sec:related works}

\textbf{Text-to-Image Generation.}
The emergence of diffusion-based text-to-image (T2I) models~\citep{nichol2021glide, saharia2022photorealistic, ramesh2022hierarchical, rombach2022high} marks a significant leap in text-to-image quality, attracting widespread attention. This paradigm has been further advanced by subsequent works~\citep{podell2023sdxl, esser2024scaling, blackforestlabs_flux}, which elevated generation quality to new heights by refining model architectures and expanding parameter capacity.

On the other hand, autoregressive T2I models~\citep{ramesh2021zero, chang2022maskgit, tian2024visual} have also been developing rapidly. Some works~\citep{team2024chameleon, xie2024show, chen2025janus, deng2025emerging} attempt to unify generation and understanding into a single model, thereby enabling the text-to-image process to better leverage vast world knowledge. As instruction following, aesthetics, and resolution have progressively improved, higher demands have been placed on generation quality, and the presence of artifacts in images has gradually become one of the main bottlenecks for improving image quality.

\textbf{Text-to-Image Evaluation.}
The evaluation of T2I models is critical for guiding their development and has evolved in tandem with their capabilities. Early metrics like Inception Score (IS)\citep{salimans2016improved} and Fréchet Inception Distance (FID)\citep{heusel2017gans} assessed general image quality, independent of text prompts. As models' instruction-following abilities improved, evaluation branched into two major directions: semantic and aesthetic alignment.

On the semantic front, metrics like CLIP Score~\citep{hessel2021clipscore} were introduced to measure text-image consistency. This was followed by specialized benchmarks targeting specific abilities, such as object-level alignment in Geneval~\citep{ghosh2023geneval}, compositional accuracy in T2I-CompBench~\citep{huang2023t2i}, and adherence to dense prompts in DPG-Bench~\citep{hu2024ella}. In parallel, the aesthetic front focused on aligning models with human preferences. This led to influential reward models trained on large-scale human feedback, including ImageReward~\citep{xu2023imagereward}, the Human Preference Score (HPS) series~\citep{wu2023human, wu2023humanv2}, and Pick-a-Pic~\citep{kirstain2023pick}, which have become standards for judging visual appeal.

More recently, research has shifted towards evaluating higher-level attributes like plausibility and realism. For instance, RichHF~\citep{liang2024rich} focuses on human feedback for plausibility, while HEIE~\citep{yang2025heie} and FakeVLM~\citep{wen2025spot} attempt to not only detect but also explain unrealistic or "fake" elements in images. Despite these advancements, existing evaluation methods lack a granular assessment of common image artifacts and artifacts. This gap highlights the need for a comprehensive benchmark dedicated to the multi-dimensional evaluation of such artifacts, which remains a key barrier to achieving true photorealism and reliability.

\textbf{GRPO in Vision Language Model.}
With advancements in reinforcement learning\cite{schulman2017proximal, rafailov2023direct, wallace2024diffusion}, the Group Relative Policy Optimization (GRPO)~\citep{shao2024deepseekmath} algorithm has effectively enhanced the reasoning capabilities of models, showing significant potential. Consequently, recent work has begun to extend GRPO to Vision-Language Models (VLMs)~\citep{zheng2025easyr1, shen2025vlm, wang-2025-open-r1-video}, demonstrating considerable performance gains in multimodal tasks. A notable application involves leveraging scores from evaluation benchmarks~\citep{liu2025flow, xue2025dancegrpo}, such as Geneval, as reward signals to fine-tune text-to-image models~\citep{liu2025flow}, leading to significant improvements on those specific metrics. To better identify artifacts in generated images, we introduce GRPO to the task of artifacts detection.

\begin{figure*}[t]
\centering
\includegraphics[width=1.0\textwidth, trim=0cm 0cm 4cm 0cm, clip]{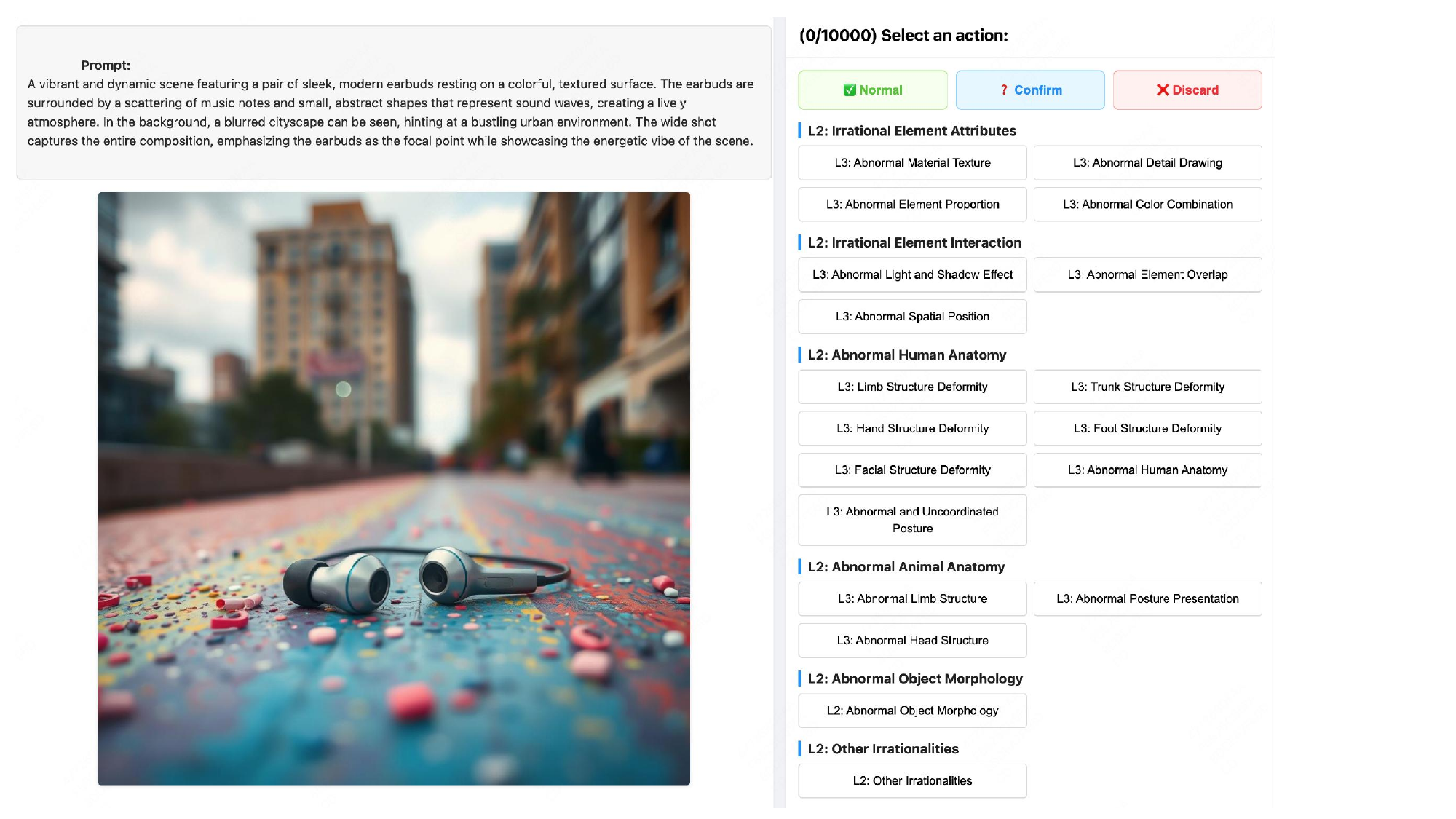} 
\caption{The annotation interface designed for the MagicData340K dataset.}
\label{fig:Label_UI}
\end{figure*}

\begin{figure*}[t]
\centering
\includegraphics[width=1.0\textwidth, trim=0cm 4cm 0cm 0cm, clip]{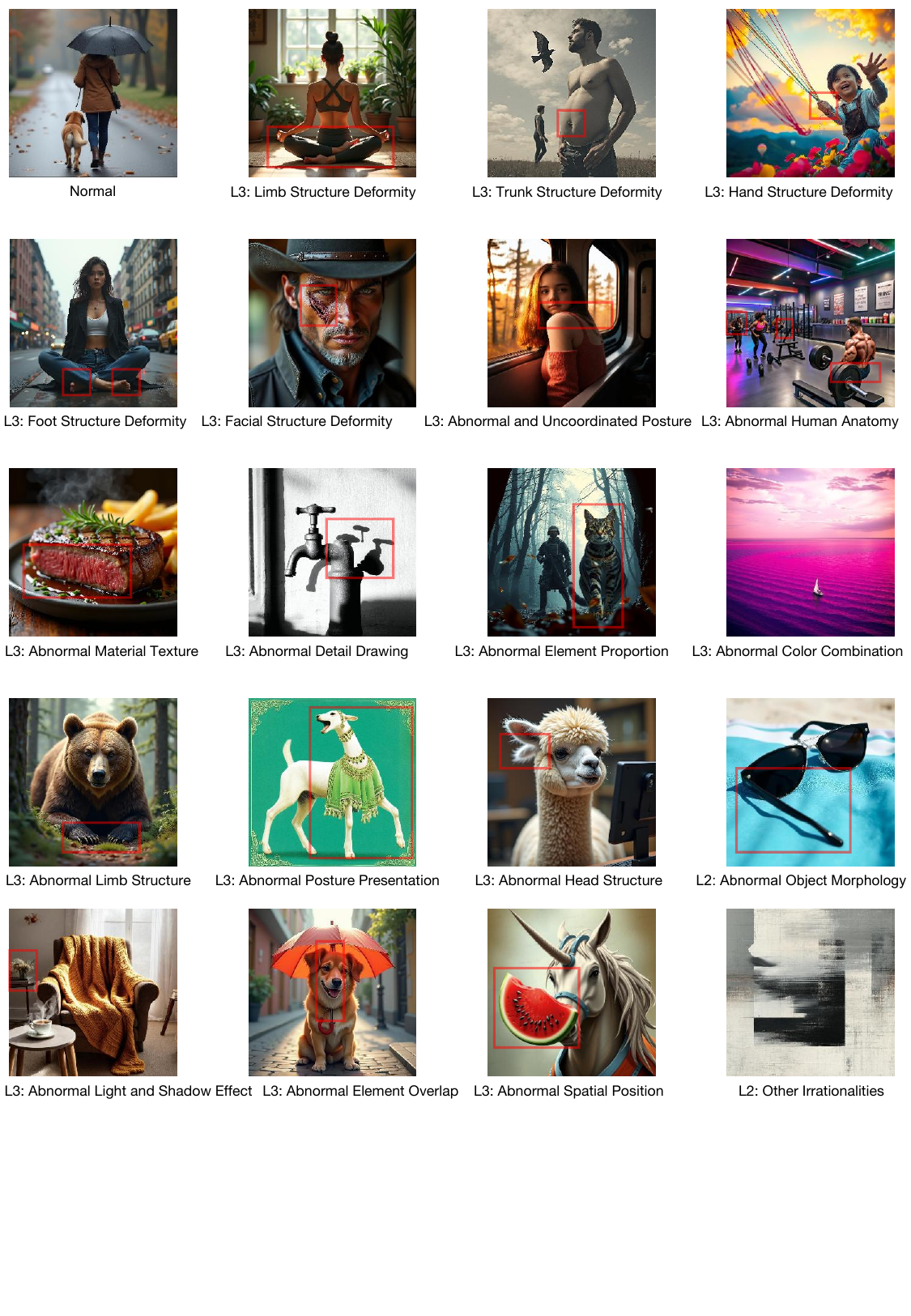} 
\caption{Visual examples of labeled artifacts categories.}
\label{fig:Label_Examples}
\end{figure*}

\begin{table}[]
\caption{Detailed statistics of the MagicData340K dataset.}
\label{table:detail_data_statistics}
\centering
\resizebox{\columnwidth}{!}{
\renewcommand{\arraystretch}{1.1}
\begin{tabular}{lll|r|c|c|c}
\hline
\multicolumn{3}{l|}{Labels}                                                                            & \multicolumn{1}{l|}{}                                                            & Train           & Test           & CoT           \\ \hline
\multicolumn{3}{l|}{Total}                                                                             & \multicolumn{1}{l|}{\textbf{(343269)}}                                           & \textbf{325238} & \textbf{17366} & \textbf{1835} \\ \hline
\multicolumn{3}{l|}{Normal}                                                                            & \multicolumn{1}{l|}{\textbf{(173768)}}                                           & \textbf{165078} & \textbf{8690}  & \textbf{642}  \\ \hline
\multicolumn{3}{l|}{Artifacts}                                                                      & \multicolumn{1}{l|}{\textbf{(169501)}}                                           & \textbf{160160} & \textbf{8676}  & \textbf{1193} \\ \hline
\multicolumn{1}{l|}{} & \multicolumn{2}{l|}{L2: Irrational Element Attributes}                         & \multicolumn{1}{l|}{\textbf{(881 / 169501) (0.52\%)}}                            & \textbf{561}    & \textbf{247}   & \textbf{256}  \\ \hline
\multicolumn{1}{l|}{} & \multicolumn{1}{l|}{}   & L3: Abnormal Material Texture                        & (54) (0.03\%)                                                                    & 37              & 19             & 37            \\ \hline
\multicolumn{1}{l|}{} & \multicolumn{1}{l|}{}   & L3: Abnormal Detail Drawing                          & (192) (0.11\%)                                                                   & 133             & 59             & 101           \\ \hline
\multicolumn{1}{l|}{} & \multicolumn{1}{l|}{}   & L3: Abnormal Element Proportion                      & (612) (0.36\%)                                                                   & 374             & 163            & 101           \\ \hline
\multicolumn{1}{l|}{} & \multicolumn{1}{l|}{}   & L3: Abnormal Color Combination                       & (23) (0.01\%)                                                                    & 17              & 9              & 17            \\ \hline
\multicolumn{1}{l|}{} & \multicolumn{2}{l|}{{\color[HTML]{00B0F0} L2: Irrational Element Interaction}} & \multicolumn{1}{l|}{{\color[HTML]{00B0F0} \textbf{(61994 / 169501) (36.6\%)}}}   & \textbf{56262}  & \textbf{3004}  & \textbf{299}  \\ \hline
\multicolumn{1}{l|}{} & \multicolumn{1}{l|}{}   & L3: Abnormal Light and Shadow Effect                 & (258) (0.15\%)                                                                   & 176             & 77             & 101           \\ \hline
\multicolumn{1}{l|}{} & \multicolumn{1}{l|}{}   & {\color[HTML]{C71C31} L3: Abnormal Element Overlap}  & {\color[HTML]{C71C31} (45857) (27.0\%)}                                          & 42387           & 2232           & 111           \\ \hline
\multicolumn{1}{l|}{} & \multicolumn{1}{l|}{}   & L3: Abnormal Spatial Position                        & (15879) (9.3\%)                                                                  & 15128           & 798            & 101           \\ \hline
\multicolumn{1}{l|}{} & \multicolumn{2}{l|}{{\color[HTML]{FF0000} L2: Abnormal Human Anatomy}}         & \multicolumn{1}{l|}{{\color[HTML]{FF0000} \textbf{(104289 / 169501) (61.53\%)}}} & \textbf{97882}  & \textbf{5645}  & \textbf{639}  \\ \hline
\multicolumn{1}{l|}{} & \multicolumn{1}{l|}{}   & L3: Limb Structure Deformity                         & (3638) (2.1\%)                                                                   & 3246            & 362            & 101           \\ \hline
\multicolumn{1}{l|}{} & \multicolumn{1}{l|}{}   & L3: Trunk Structure Deformity                        & (723) (0.1\%)                                                                    & 497             & 215            & 101           \\ \hline
\multicolumn{1}{l|}{} & \multicolumn{1}{l|}{}   & {\color[HTML]{C71C31} L3: Hand Structure Deformity}  & {\color[HTML]{C71C31} (54114) (31.9\%)}                                          & 50959           & 2684           & 269           \\ \hline
\multicolumn{1}{l|}{} & \multicolumn{1}{l|}{}   & L3: Foot Structure Deformity                         & (4604) (2.7\%)                                                                   & 4113            & 459            & 101           \\ \hline
\multicolumn{1}{l|}{} & \multicolumn{1}{l|}{}   & L3: Facial Structure Deformity                       & (8581) (5.0\%)                                                                   & 7661            & 853            & 101           \\ \hline
\multicolumn{1}{l|}{} & \multicolumn{1}{l|}{}   & {\color[HTML]{C71C31} L3: Abnormal Human Anatomy}    & {\color[HTML]{C71C31} (42327) (24.9\%)}                                          & 40015           & 2108           & 104           \\ \hline
\multicolumn{1}{l|}{} & \multicolumn{1}{l|}{}   & L3: Abnormal and Uncoordinated Posture               & (300) (0.1\%)                                                                    & 207             & 91             & 101           \\ \hline
\multicolumn{1}{l|}{} & \multicolumn{2}{l|}{L2: Abnormal Animal Anatomy}                               & \multicolumn{1}{l|}{\textbf{(23464/169501) (13.84\%)}}                           & \textbf{21968}  & \textbf{1324}  & \textbf{235}  \\ \hline
\multicolumn{1}{l|}{} & \multicolumn{1}{l|}{}   & {\color[HTML]{C71C31} L3: Abnormal Limb Structure}   & {\color[HTML]{C71C31} (20932) (12.3\%)}                                          & 19760           & 1041           & 130           \\ \hline
\multicolumn{1}{l|}{} & \multicolumn{1}{l|}{}   & L3: Abnormal Posture Presentation                    & (234) (0.1\%)                                                                    & 162             & 72             & 101           \\ \hline
\multicolumn{1}{l|}{} & \multicolumn{1}{l|}{}   & L3: Abnormal Head Structure                          & (5433) (3.2\%)                                                                   & 4833            & 539            & 101           \\ \hline
\multicolumn{1}{l|}{} & \multicolumn{2}{l|}{L2: Abnormal Object Morphology}                            & \multicolumn{1}{l|}{\textbf{(37085 / 169501) (21.88\%)}}                         & \textbf{34972}  & \textbf{1842}  & \textbf{101}  \\ \hline
\multicolumn{1}{l|}{} & \multicolumn{2}{l|}{L2: Other Irrationalities}                                 & \multicolumn{1}{l|}{\textbf{(20) (0.01\%)}}                                      & \textbf{18}     & \textbf{10}    & \textbf{18}   \\ \hline
\end{tabular}
}
\end{table}

\subsection{Dataset Annotation and Label Taxonomy}
\label{ap-sec: dataset and label}

This section provides a comprehensive overview of the MagicData340K dataset, detailing the data statistics, the established labeling methodology, and the complete taxonomy of our classification labels.

Our dataset contains a total of 343,269 images, which are partitioned into a training set (325,238), a test set (17,366), and a Chain-of-Thought (CoT) set (1,294). The data is broadly categorized into ``Normal" images (173,768) and ``Artifacts" images (169,501), which contain various types of anomalies. A detailed statistical breakdown is presented in Table~\ref{table:detail_data_statistics}. The analysis reveals that ``Abnormal Human Anatomy" is the most prevalent category of artifacts, accounting for 61.53\% of all anomalous samples. This is followed by ``Irrational Element Interaction" (36.6\%) and ``Abnormal Object Morphology" (21.88\%), highlighting the most common failure modes in image generation.

To ensure consistency and quality in annotation, we develop a specific labeling user interface (UI), as shown in Fig.~\ref{fig:Label_UI}, and establish a set of core guidelines for our annotators. These guidelines are as follows: 
\begin{enumerate}
    \item \textbf{Handling Multiple Issues:} When judging, if an image corresponds to multiple issues, only the two most obvious issues need to be marked. However, there are two exceptions: if the number of people in the image is $\geq$3 and the number of abnormal issues is $\geq$3, you can simply label it as ``L3: Abnormal Human Anatomy"; if a single person has more than 3 abnormal issues, you can directly label it as ``L3: Abnormal Human Anatomy".
    \item \textbf{Standard for Abnormality:} During the annotation process, only more obvious abnormalities need to be noted. If an issue cannot be identified within 3 seconds (e.g., it may be a blurred background entity or a small-sized object), or if a reasonable explanation exists, or it belongs to an imaginable special case, it is not considered abnormal; you can label it as ``Whether Normal: True".
    \item \textbf{Distinguishing Style from Structure:} Instantly fake or heavily AI-generated feeling belongs to image style issues, not structural abnormalities; therefore, you can label it as ``Whether Normal: True".
    \item \textbf{Ignoring Textual Content:} If there is a text error in the image, do not judge it as unreasonable based on the text content.
    \item \textbf{Considering User Prompts:} When some images initially appear to violate physical laws, refer to the user's prompt and style requirements to assist judgment. If the image conforms to the special conditions specified in the prompt, you can label it as ``Whether Normal: True".
\end{enumerate}

The artifacts were categorized into a hierarchical taxonomy with five high-level (L2) categories, each containing more specific (L3) sub-categories. Fig.~\ref{fig:Label_Examples} provides visual examples for many of these labels. The complete definitions are as follows:

L2: Irrational Element Attributes: ``The visual attributes of elements in the image do not conform to physical laws."
\begin{itemize}
	\item L3: Abnormal Material Texture: ``The material texture does not match the actual properties of the object, such as metallic texture displaying a wooden pattern."
	\item L3: Abnormal Detail Drawing: ``Abnormal background elements in the image."
	\item L3: Abnormal Element Proportion: ``The relative sizes of elements in the image do not conform to real proportions or expected scales, such as a mosquito larger than a hand."
	\item L3: Abnormal Color Combination: ``Color combination violates visual color theory, leading to a visual appearance that does not conform to the real world."
\end{itemize}

L2: Irrational Element Interaction: ``The spatial and logical interactions between elements in the image are unreasonable."
\begin{itemize}
	\item L3: Abnormal Light and Shadow Effect: ``The position of the light source and shadow direction are inconsistent, causing unnatural light and shadow projection. The light and shadow effect does not match the light source position, intensity, and objective factors."
	\item L3: Abnormal Element Overlap: ``Overlap relationships between different elements do not conform to physical laws, such as a solid object partially penetrating another object."
	\item L3: Abnormal Spatial Position: ``The distribution and logical arrangement of elements in space are inconsistent, causing chaotic overall layout, such as floating, mismatch between inside and outside state in a mirror."
\end{itemize}

L2: Abnormal Human Anatomy: ``The structure of the human body in the image does not conform to normal physiological and anatomical standards"
\begin{itemize}
	\item L3: Limb Structure Deformity: ``Limb structure does not conform to conventional human form."
	\item L3: Trunk Structure Deformity: ``The spine shows unnatural curvature or twisting."
	\item L3: Hand Structure Deformity: ``Abnormal number of fingers or unreasonable joint angles."
	\item L3: Foot Structure Deformity: ``Disorganized toe arrangement or abnormal arch shape."
	\item L3: Facial Structure Deformity: ``Imbalance in facial features or lack of facial symmetry."
	\item L3: Abnormal Human Anatomy: ``Multiple human abnormalities."
	\item L3: Abnormal and Uncoordinated Posture: ``Whole body posture does not conform to gravitational direction or movements are inconsistent with ergonomics."
\end{itemize}

L2: Abnormal Animal Anatomy: ``The structure of animals in the image does not conform to normal physiological and anatomical standards"
\begin{itemize}
	\item L3: Abnormal Limb Structure: ``Imbalance in animal limb proportions or shape does not conform to common sense."
	\item L3: Abnormal Posture Presentation: ``Animal movement posture does not match its biological characteristics."
	\item L3: Abnormal Head Structure: ``Abnormal position or imbalance in the proportion of eyes or ears."
\end{itemize}

L2: Abnormal Object Morphology: ``Geometric shape is abnormal, the object outline or geometric proportions do not match actual characteristics; or the construction is unreasonable, the connection method of object parts does not conform to logic or actual structure."

L2: Other Irrationalities: ``Other irrationalities."

\subsection{Chatting Template}
\label{ap-sec:prompt design}

To effectively guide the model in identifying and classifying anomalies in generated images, we designed a comprehensive prompt structured with the following key components:
\begin{enumerate}

    \item \textbf{Task Description:} The core objective is then clearly defined: ``You need to determine whether this image is reasonable (or whether there is any deformity), and if it is not reasonable, provide the corresponding type of deformity. If the provided type of deformity has sub-tags, additionally provide the corresponding sub-tag categories."
    
    \item \textbf{Labels and Definitions:} To ensure standardized classification, a complete taxonomy of deformities is provided: ``All types of deformities and their sub-tags are: `` + str(type\_definition) + ". Note that the primary label is Whether Normal, L2 represents second-level tags, and L3 represents third-level tags."
    
    \item \textbf{Illustrative Answer Formats:} To clarify the expected output, the prompt provides a few example answer formats. These examples demonstrate the required structure for different scenarios, such as 
    \begin{itemize}
        \item the format for a normal image (\{``Whether Normal": True\})
        \item an image with an abnormality but no sub-tags (\{``Whether Normal": False, ``Type of Abnormality": \{``L2: Abnormal Object Morphology": True\}\})
        \item an image with an abnormality that includes sub-tags (\{``Whether Normal": False, ``Type of Abnormality": \{``L2: Abnormal Human Anatomy": [``L3: Abnormal Human Anatomy"]\}\})
        \item an image with two distinct types of abnormalities (\{``Whether Normal": False, ``Type of Abnormality": \{``L2: Abnormal Object Morphology": True, ``L2: Abnormal Human Anatomy": [``L3: Abnormal Human Anatomy"]\}\}).
    \end{itemize}
    
    \item \textbf{Answering Process:} The prompt outlines the required cognitive steps for the model: ``You need to first understand all the given labels and rules, then think about possible issues according to the text prompt and the subject of the prompt, and then observe the image to analyze every detail in the image to determine whether there is any deformity."
    
    \item \textbf{Content Requirements:} The model is instructed on how to articulate its reasoning process: ``Give a continuous thinking process using natural language. The response should flow seamlessly as a narrative or story, examining the image as a whole rather than in separate points. Please describe the reasoning process without using bullet points or distinct sections."
    
    \item \textbf{Formatting Requirements:} Finally, the required output structure is specified: ``Ensure that the answer matches the format of the given example. The output format should be $<think>...</think>...\ boxed\{answer\}$."
\end{enumerate}

\subsection{Multi-Reward Design}
\label{ap-sec: multi-reward}

Consistency Reward ($r_c$): To prevent reward hacking, where the model might produce a correct final answer without a valid reasoning process, we use a smaller, pre-trained LLM to judge the logical consistency between the generated Chain-of-Thought and the final labeled output. The result is a binary reward, $r_c \in \{0,1\}$, where 1 indicates consistency. This entire reward is nullified if the reasoning is flawed.

Hierarchical Rewards ($r_1,r_2,r_3,r_4$):
The system categorizes rewards into four hierarchical levels:
\begin{enumerate}
    \item $r_0$ (Format and Parsability Reward): This is a binary reward, $r_0 \in \{0,1\}$. It assesses whether the output adheres to the required $<think>$ and dictionary format and can be successfully parsed by our evaluation script. This includes checking for hierarchical dependencies (e.g., an L3 label only appears under its parent L2 label).
    \item $r_1$ (L1-level Label Reward): This is a binary reward, $r_1 \in \{0,1\}$, evaluating the accuracy of the highest-level binary classification (i.e., whether the image is correctly identified as "Normal" or "Artifact").
    \item $r_2,r_3$ (L2, L3-level Label Rewards): Since a single sample can have multiple L2 and L3 labels, these rewards are calculated based on the number of correct ($n_{correct}$), missed ($n_{miss}$), and extra ($n_{extra}$) predictions. The score is 1 if all ground-truth labels are correctly predicted and there are no extra predictions, and 0 if no labels are predicted at all. Otherwise, the score is calculated using the following formula, which rewards precision and penalizes recall errors:
\begin{equation}
r_{2,3} = \text{clamp}(0.6 \cdot n_{correct} - 0.3 \cdot (n_{miss} + n_{extra}), \min=0, \max=1).
\end{equation}
\end{enumerate}

\subsection{Experimental Details}
\label{ap-sec:Experimental Details}
We provide supplementary information for our experiments. Section~\ref{ap-sec:Experimental Details} details the experimental setup and the formulas used to calculate our evaluation metrics. Furthermore, Table~\ref{table:ablation 4 class} presents the detailed per-class results of the ablation study discussed in the main text.

\subsubsection{Experimental Setup}

All of our experiments are built upon the \textbf{Qwen2.5-VL-7B} as the foundational model. Our training methodology follows a two-stage process designed to progressively enhance the model's capabilities. The initial cold start phase consists of Supervised Fine-Tuning (SFT) using our curated \textbf{CoT dataset}, which contains 1,294 high-quality examples with detailed reasoning. This stage was conducted for 40 steps with a batch size of 32, a learning rate of 2.0e-5, and a LoRA rank of 32. Following this, the model undergoes Group Relative Policy Optimization (GRPO) for 300 steps on our main \textbf{GRPO dataset}, which is partitioned into 325,238 training and 17,366 test samples. For the GRPO stage, we used the AdamW optimizer with a learning rate of 1.0e-6 and a global batch size of 128. To benchmark the effectiveness of this two-stage approach, we also trained an SFT-only baseline model on the full dataset for 300 steps.

\subsubsection{Evaluation Metrics}
To quantitatively assess model performance, we employ the standard metrics of Precision, Recall, and F1-Score. Precision measures the accuracy of positive predictions, while Recall measures the model's ability to identify all relevant instances. The F1-Score provides a balanced measure as the harmonic mean of Precision and Recall. The formulas are as follows:

\begin{equation}
    \text{Precision} = \frac{\text{TP}}{\text{TP} + \text{FP}}
\end{equation}
\begin{equation}
    \text{Recall} = \frac{\text{TP}}{\text{TP} + \text{FN}}
\end{equation}
\begin{equation}
    \text{F1-Score} = 2 \cdot \frac{\text{Precision} \cdot \text{Recall}}{\text{Precision} + \text{Recall}}
\end{equation}

where:
\begin{itemize}
    \item \textbf{TP (True Positives):} The number of images correctly identified as having a specific artifacts (model predicted the label, and it was correct).
    \item \textbf{FP (False Positives):} The number of images incorrectly identified as having a specific artifacts when they do not (model \textbf{over-predicted} the label).
    \item \textbf{FN (False Negatives):} The number of images that have a specific artifacts but were not identified by the model (model \textbf{under-predicted} the label).
\end{itemize}

Our evaluation is conducted in two settings: a top-level binary detection of ``Normal" vs. ``Artifact" images, and a fine-grained, multi-label classification across our four primary L2 artifacts categories. For the multi-label task, we report both macro and micro averages to provide a complete performance picture. The \textbf{Macro Average} computes the metric independently for each class and then takes the average, treating all classes equally. The \textbf{Micro Average} aggregates the contributions of all classes to compute the metric globally, giving more weight to more populous classes. The formulas for a set of C classes are:

\begin{equation}
    \text{Macro Average} = \frac{1}{C} \sum_{i=1}^{C} \text{Metric}_i
\end{equation}

\begin{equation}
    \text{Micro Average Precision} = \frac{\sum_{i=1}^{C} \text{TP}_i}{\sum_{i=1}^{C} (\text{TP}_i + \text{FP}_i)}, \quad \text{Micro Average Recall} = \frac{\sum_{i=1}^{C} \text{TP}_i}{\sum_{i=1}^{C} (\text{TP}_i + \text{FN}_i)}
\end{equation}

\begin{table}[]
\caption{Four class performance of ablation study.}
\label{table:ablation 4 class}
\centering
\resizebox{\columnwidth}{!}{
\renewcommand{\arraystretch}{1.2}
\begin{tabular}{cccccccccccccc}
\hline
                                           &                            & \multicolumn{3}{c}{L2: Irrational Element Interaction} & \multicolumn{3}{c}{L2: Abnormal Human Anatomy}      & \multicolumn{3}{c}{L2: Abnormal Animal Anatomy}     & \multicolumn{3}{c}{L2: Abnormal Object Morphology}  \\ \cmidrule(lr){3-5} \cmidrule(lr){6-8} \cmidrule(lr){9-11} \cmidrule(lr){12-14} 
                                           &                            & Precision        & Recall           & F1               & Precision       & Recall          & F1              & Precision       & Recall          & F1              & Precision       & Recall          & F1              \\ \hline
\multicolumn{1}{c|}{\multirow{4}{*}{GRPO}} & MagicAssessor-7B           & \textbf{0.3665}  & 0.4621           & 0.4088           & 0.8957          & 0.6177          & 0.7312          & 0.5805          & 0.6563          & \underline{0.6161}    & \textbf{0.3359} & 0.3616          & \textbf{0.3482} \\
\multicolumn{1}{c|}{}                      & w/o Multi-Bucket Sampling         & 0.2248           & 0.0097           & 0.0185           & 0.7407          & \textbf{0.8955} & \textbf{0.8107} & 0.4621          & \textbf{0.9071} & 0.6123          & 0.1468          & \textbf{0.5233} & 0.2293          \\
\multicolumn{1}{c|}{}                      & w/o Positive Data Resampling & \underline{0.3575}     & \underline{0.4917}     & \textbf{0.4140}  & \textbf{0.9010} & 0.5624          & 0.6926          & \textbf{0.5949} & 0.6322          & 0.6130          & \underline{0.3315}    & 0.3529          & \underline{0.3418}    \\
\multicolumn{1}{c|}{}                      & w/o Consistent Reward      & 0.3258           & \textbf{0.5523}  & \underline{0.4098}     & \underline{0.8980}    & 0.5488          & 0.6813          & \underline{0.5834}    & \underline{0.6579}    & \textbf{0.6184} & 0.3096          & \underline{0.3735}    & 0.3386          \\ \hline
\multicolumn{2}{c}{SFT}                                                & 0.2879           & 0.4567           & 0.3532           & 0.7640          & \underline{0.7382}    & \underline{0.7509}    & 0.5509          & 0.4615          & 0.5023          & 0.2124          & 0.0798          & 0.1160         \\ \hline
\end{tabular}
}
\end{table}

\begin{figure*}[t]
\centering
\includegraphics[width=1.0\textwidth, trim=0cm 0cm 1cm 0cm, clip]{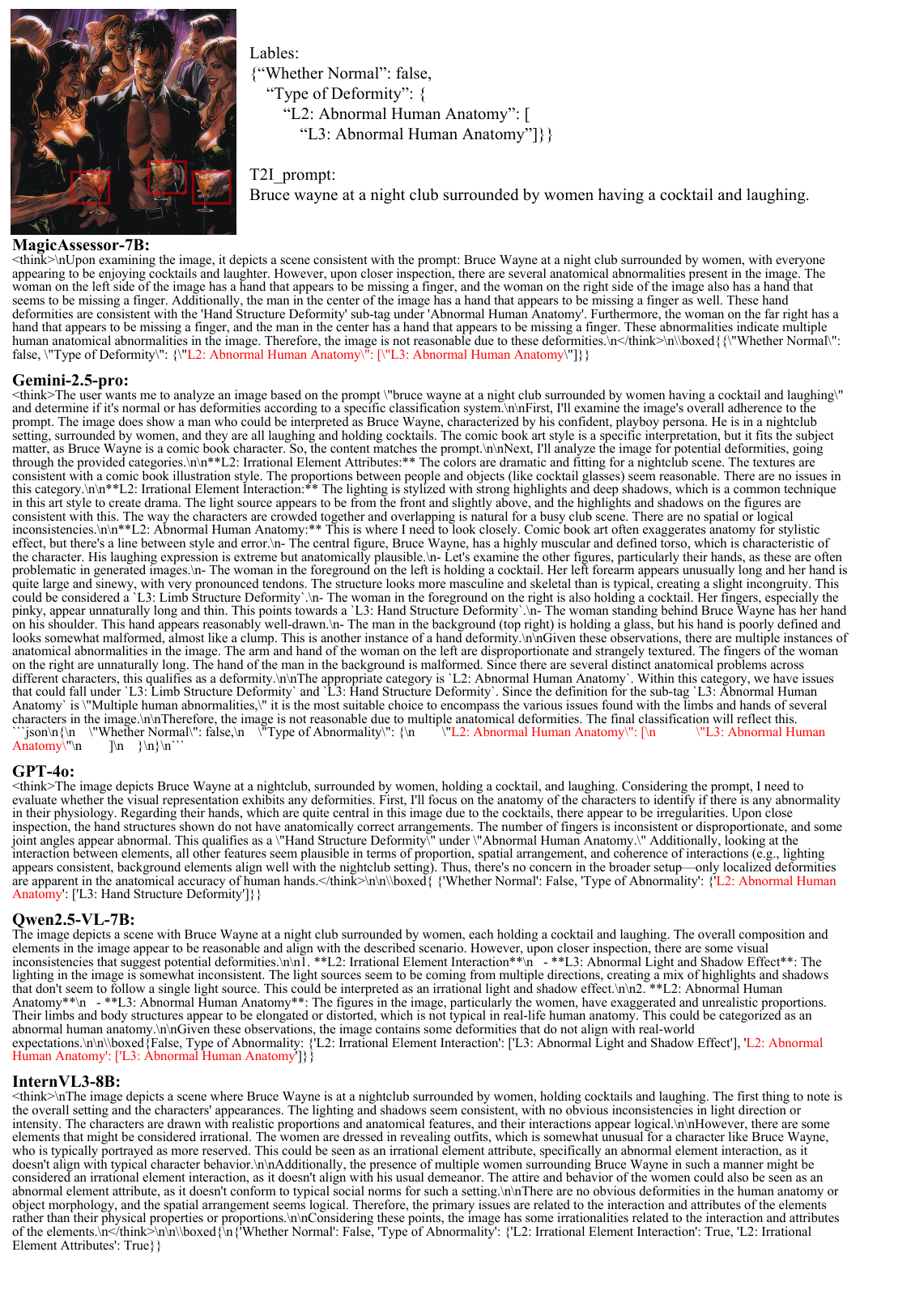} 
\caption{Model outputs for label ``L2: Abnormal Human Anatomy".}
\label{fig:model_response_human}
\end{figure*}

\begin{figure*}[t]
\centering
\includegraphics[width=1.0\textwidth, trim=0cm 0cm 1cm 0cm, clip]{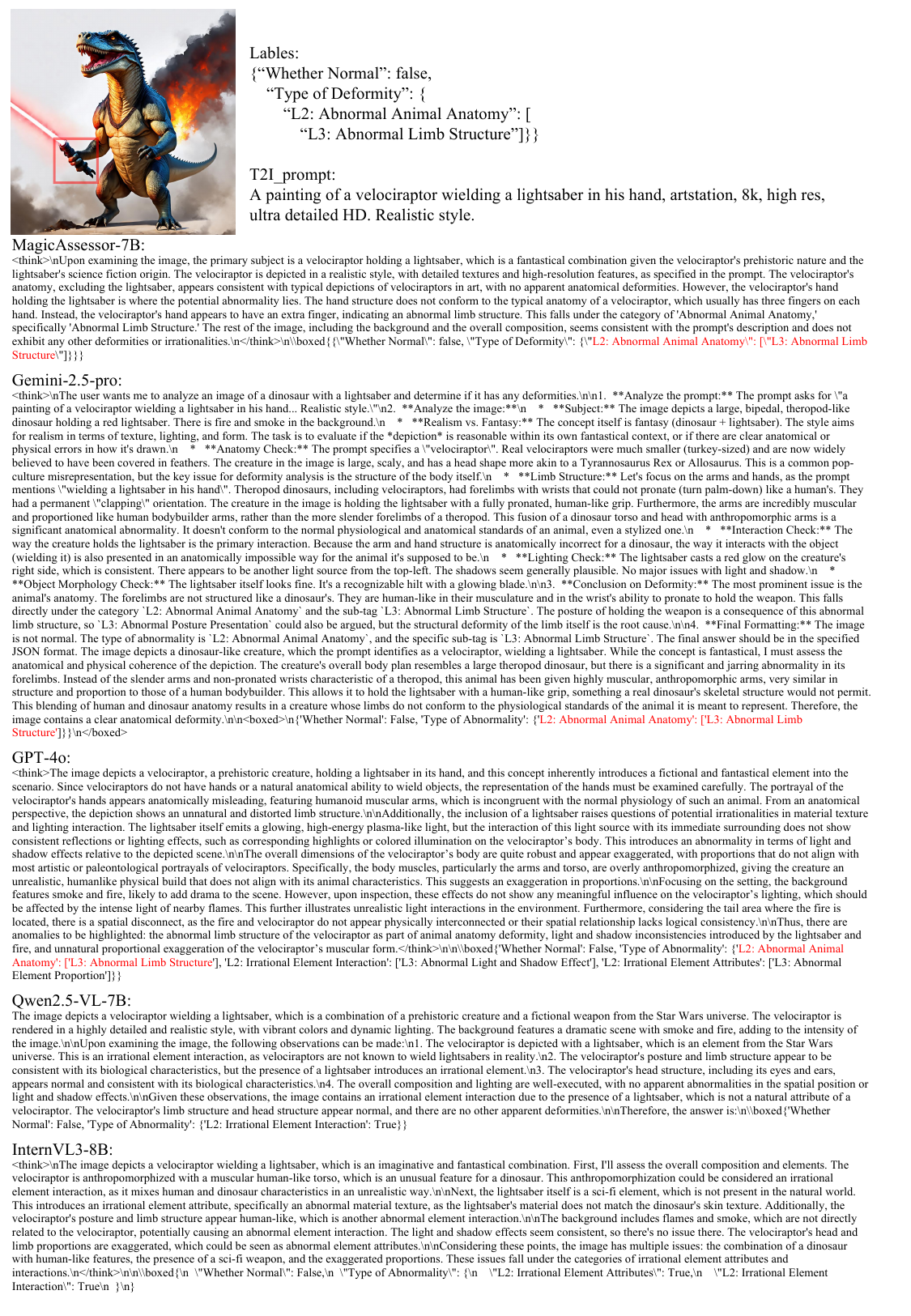} 
\caption{Model outputs for label ``L2: Abnormal Animal Anatomy".}
\label{fig:model_response_animal}
\end{figure*}

\begin{figure*}[t]
\centering
\includegraphics[width=1.0\textwidth, trim=0cm 0cm 1cm 0cm, clip]{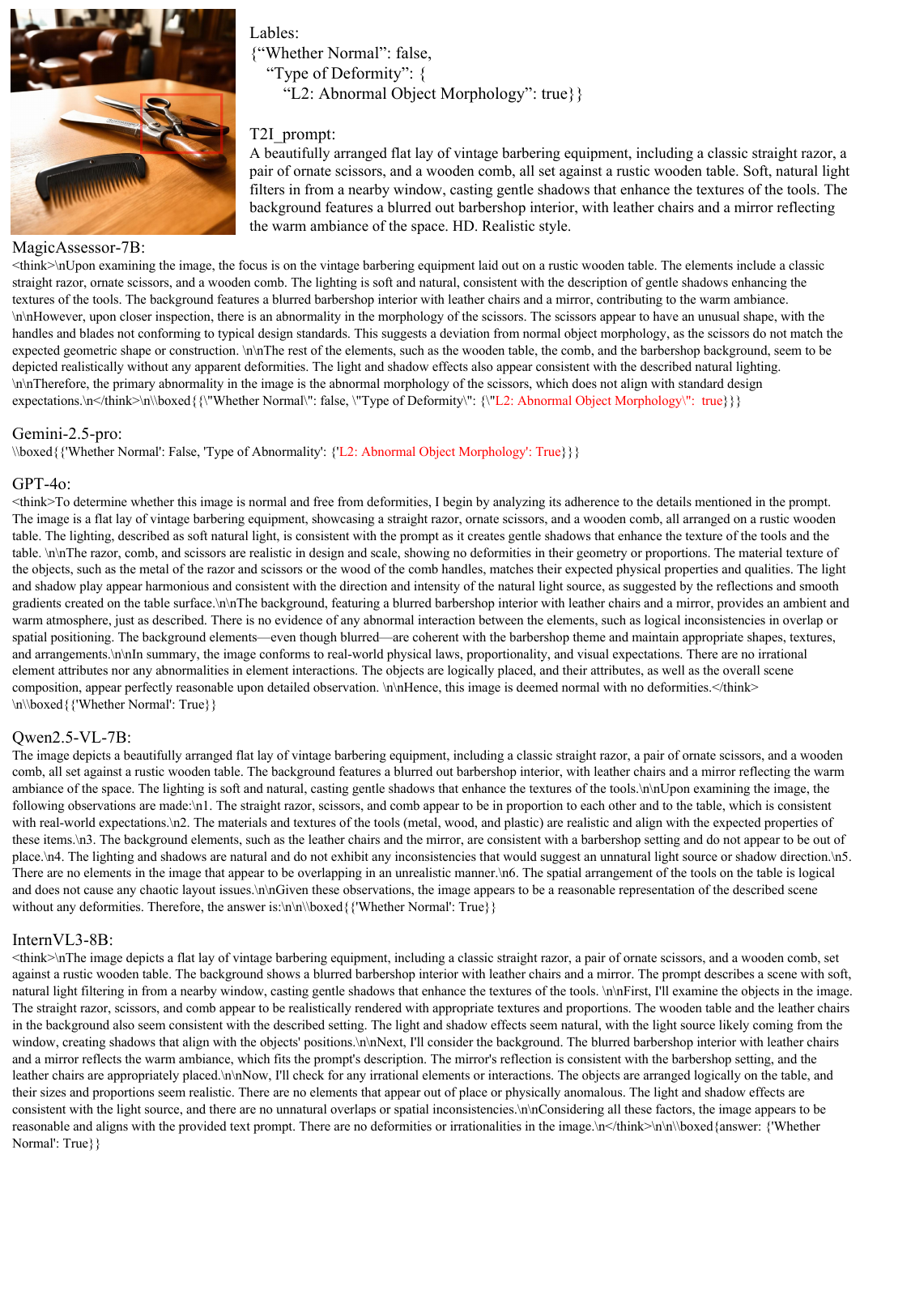} 
\caption{Model outputs for label ``L2: Abnormal Object Morphology".}
\label{fig:model_response_object}
\end{figure*}

\begin{figure*}[t]
\centering
\includegraphics[width=1.0\textwidth, trim=0cm 0cm 1cm 0cm, clip]{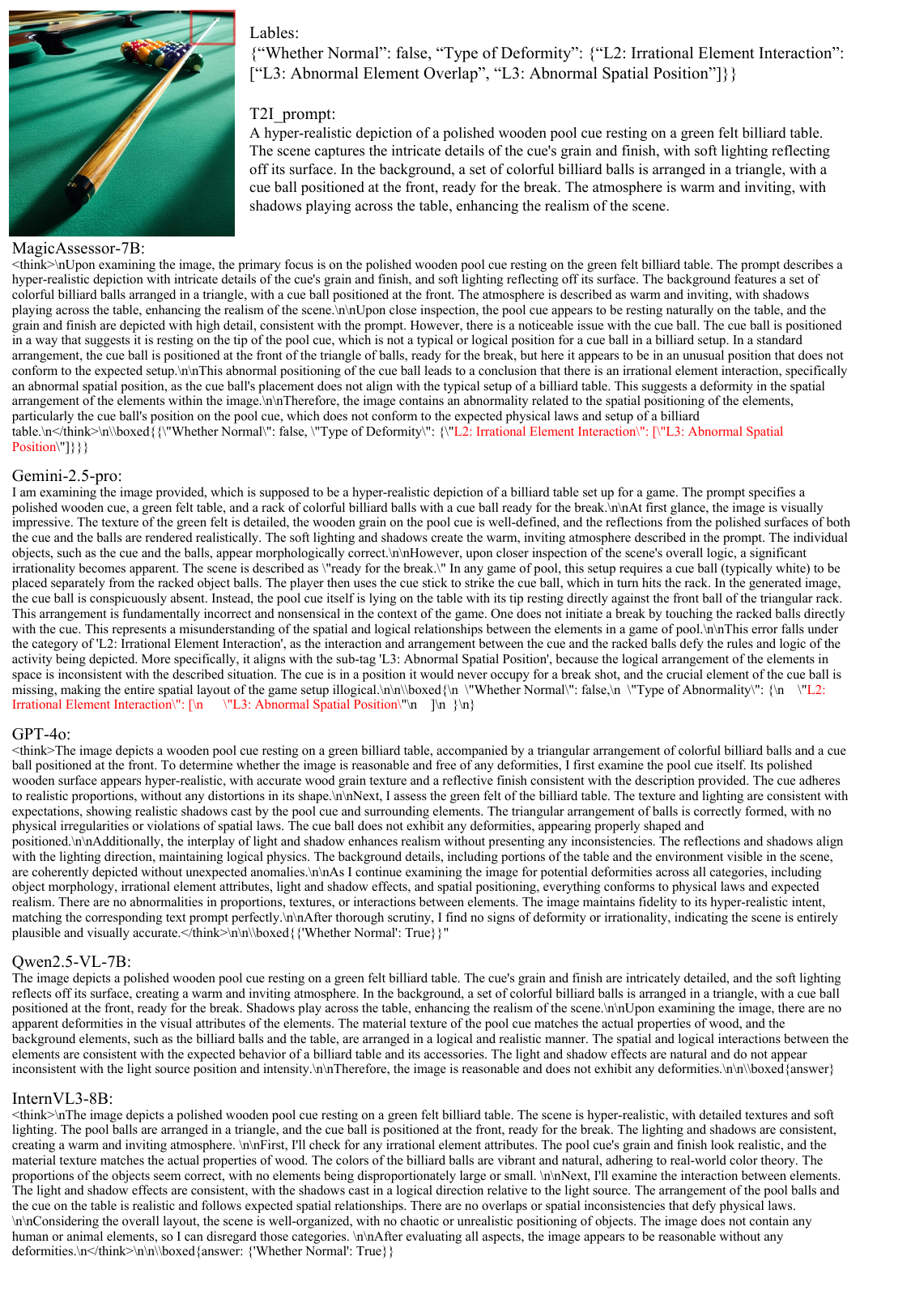} 
\caption{Model outputs for label ``L2: Irrational Element Interaction".}
\label{fig:model_response_interaction}
\end{figure*}

\subsection{Response of Different Models}
\label{ap-sec:Response of Different Models}
We selected two representative cases to showcase the distinct answering styles and behavioral patterns of the various models, as shown in Fig.~\ref{fig:model_response_human}, Fig.~\ref{fig:model_response_animal}, Fig.~\ref{fig:model_response_object} and Fig.~\ref{fig:model_response_interaction}.

\begin{itemize}
    \item MagicAssessor-7B demonstrates a highly focused and efficient response style. In all four examples—from animal anatomy and human figures to object interaction and morphology—its reasoning process was concise and directly targeted the specific visual flaw. It consistently avoided irrelevant descriptions, instead moving straight to identifying and correctly classifying the error. This pattern highlights its reliable and accurate analytical approach.
    \item Gemini-2.5-pro is characterized by its thorough and highly accurate analytical process. It correctly identified the abnormalities in all cases, providing detailed, step-by-step reasoning that demonstrates a deep contextual understanding. For instance, in the billiard table scene, it not only spotted the error but also explained why it was illogical according to the rules of the game. Its extensive reasoning serves to strengthen its conclusions, rather than leading it to incorrect classifications.
    \item GPT-4o shows inconsistent performance. While it can identify more obvious anatomical deformities, such as the incorrect limb structure on the dinosaur and the artifact hands in the nightclub scene, it fails on more subtle or logical errors. In the billiard table and barber tool examples, it overlooked clear flaws in object arrangement and shape, incorrectly judging the images to be normal. Its tendency to provide a general description can prevent it from noticing critical details, making it unreliable.
    \item Qwen2.5-VL-7B and InternVL3-8B exhibited significant weaknesses in their analytical capabilities. They consistently failed to identify the actual deformities in the images. A common issue was misinterpreting the creative intent of a prompt as a flaw (e.g., labeling the concept of a dinosaur with a lightsaber as an error). Furthermore, they often made subjective judgments about the scene's content rather than performing an objective analysis of the generated image. Both models also frequently failed to follow the required output format.
\end{itemize}

\begin{table}[]
\caption{Evaluation score ($\uparrow$) on different labels of different models in MagicBench.}
\label{table:benchmark_detail}
\centering
\resizebox{0.85\columnwidth}{!}{
\begin{tabular}{ccccccccccc}
\hline
       & \multicolumn{3}{c}{Human}                                                     & \multicolumn{2}{c}{Animal}                       & \multicolumn{3}{c}{Object}                                                     \\ \cmidrule(lr){2-4} \cmidrule(lr){5-6} \cmidrule(lr){7-9} 
                     & Single               & \multicolumn{1}{c}{Double} & \multicolumn{1}{c}{Multiple} & Single               & \multicolumn{1}{c}{Multi} & Single               & \multicolumn{1}{c}{Multiple} & \multicolumn{1}{c}{Compose} \\ \hline
FLUX.1-dev      & 80.00        & 43.00        & 16.00        & 85.00        & 11.00        & 97.00        & 86.87        & 84.85       \\
Seedream3.0     & 93.00        & 26.00        & 4.00         & 89.00        & 5.00         & 98.99        & 83.70        & 88.42       \\
Qwen-image      & 91.00        & 34.00        & 8.08         & 83.00        & 7.07         & 98.00        & 82.98        & 81.91       \\
Hidream-l1      & 77.00        & 33.00        & 8.00         & 87.00        & 4.00         & 100.00       & 81.91        & 87.88       \\
FLUX.1-schnell  & 78.00        & 35.00        & 3.00         & 77.00        & 6.00         & 98.99        & 79.38        & 81.00       \\
SD3.5           & 78.00        & 34.00        & 8.00         & 74.00        & 10.00        & 93.94        & 80.85        & 72.63       \\
Kolors1.0       & 66.00        & 34.00        & 19.19        & 76.00        & 13.00        & 84.21        & 75.00        & 67.02       \\
SD3             & 66.00        & 15.00        & 1.00         & 83.00        & 7.00         & 97.98        & 77.17        & 64.89       \\
SDXL            & 69.00        & 21.00        & 8.00         & 83.00        & 6.00         & 92.71        & 70.21        & 59.57       \\ \hline
GPT-image-1     & 84.00        & 44.00        & 9.00         & 96.00        & 9.00         & 99.00        & 85.26        & 89.58       \\
Bagel           & 72.00        & 36.00        & 15.00        & 85.00        & 20.00        & 91.00        & 83.00        & 89.80       \\
Blip3-o         & 83.00        & 48.00        & 9.00         & 83.00        & 9.00         & 95.83        & 78.72        & 68.82       \\
Janus-pro       & 55.00        & 10.00        & 1.01         & 73.00        & 3.03         & 98.99        & 72.92        & 69.57       \\
Show-o          & 51.00        & 14.00        & 7.00         & 71.00        & 3.00         & 95.92        & 70.21        & 61.96       \\ \hline
\end{tabular}
}
\end{table}

\subsection{Prompt Construction of MagicBench}
\label{ap-sec:Prompt Construction of MagicBench}

To ensure the diversity and comprehensiveness of our evaluation, we devise a structured approach for prompt construction centered on three core entity types: Human, Animal, and Object. For each entity, we design prompts of increasing complexity to systematically test model capabilities. These are classified into eight distinct sub-categories: human\_single, human\_double, human\_multiple, animal\_single, animal\_multiple, object\_single, object\_multiple, and object\_compose. To generate a rich and varied set of prompts, we first curate an extensive entity library extracted from large-scale datasets. Subsequently, we utilize Gemini-2.5-pro~\citep{gemini2024report} to systematically combine these entities with a wide range of contexts, actions, and stylistic elements (e.g., scenes, camera shots). The detailed design for each entity type is outlined below.

\begin{itemize}
    \item For the human\_single sub-category, prompts focus on the detailed description of individuals, specifying their profession, actions, and emotional state. The human\_double sub-category assesses the capacity to represent interactions between two people, such as collaboration or conversation, testing the model's grasp of body language and composition. Finally, human\_multiple challenges the ability to render complex group scenes, such as team sports or crowded markets, to evaluate performance in managing busy compositions and generating a specific group atmosphere.
    \item The animal\_single prompts are centered on capturing the distinctive features and characteristic actions of a single animal, often styled after wildlife photography to test for anatomical accuracy and environmental realism. The animal\_multiple sub-category evaluates the capacity to depict collective animal behaviors, such as migration or herding. These prompts often employ wide-angle or aerial perspectives to test the model's ability to generate large-scale scenes while maintaining the integrity of individual animals.
    \item The object\_single prompts concentrate on the careful depiction of a single object, aiming to cover different material properties, such as the transparency of glass or the aged surface of metal, under specific lighting. Object\_multi assesses performance in generating patterns and textures from many identical objects, testing the ability to handle repetition and convey a sense of scale. Lastly, object\_compose includes rendering the logical and spatial relationships between different but thematically linked objects, testing its ability to create a coherent and well-composed scene.
\end{itemize}

After obtaining these initial prompts, we extract the corresponding subject (e.g., little boy) from each one and annotate it with its class (e.g., human) and detailed sub-category (e.g., human\_single). Finally, we append the instruction, ``The image must include the complete `subject'." to the end of each prompt to form the final test instance, where `subject' refers to the extracted subject. The scores of different models for each category are presented in Table~\ref{table:benchmark_detail}. Note that due to some generation failures, the denominator for some categories is 99 or 98, instead of 100.

\end{document}